\documentclass[twoside,11pt]{article}
\usepackage{jair, rawfonts}

\usepackage[colorlinks,urlcolor=blue,linkcolor=blue,citecolor=blue]{hyperref}
\usepackage{color,array}
\usepackage{graphicx}
\usepackage{scalerel}
\usepackage{tikz}
\usetikzlibrary{svg.path}
\usepackage{xcolor}
\usepackage{amsmath}
\DeclareMathOperator*{\argmax}{argmax}
\usepackage{amsfonts}
\usepackage{mathtools}
\usepackage[numbers]{natbib}
\usepackage[ruled,vlined,noend,linesnumbered]{algorithm2e}
\DontPrintSemicolon
\usepackage{multirow}
\usepackage{multicol}
\usepackage{booktabs}
\usepackage{blkarray}
\usepackage{fancyhdr}
\usepackage{float}
\usepackage[verbose]{placeins} 
\usepackage{wrapfig}
\usepackage{subcaption}

\definecolor{orcidlogocol}{HTML}{A6CE39}
\tikzset{
    orcidlogo/.pic={
        \fill[orcidlogocol] svg{M256,128c0,70.7-57.3,128-128,128C57.3,256,0,198.7,0,128C0,57.3,57.3,0,128,0C198.7,0,256,57.3,256,128z};
        \fill[white] svg{M86.3,186.2H70.9V79.1h15.4v48.4V186.2z}
        svg{M108.9,79.1h41.6c39.6,0,57,28.3,57,53.6c0,27.5-21.5,53.6-56.8,53.6h-41.8V79.1z M124.3,172.4h24.5c34.9,0,42.9-26.5,42.9-39.7c0-21.5-13.7-39.7-43.7-39.7h-23.7V172.4z}
        svg{M88.7,56.8c0,5.5-4.5,10.1-10.1,10.1c-5.6,0-10.1-4.6-10.1-10.1c0-5.6,4.5-10.1,10.1-10.1C84.2,46.7,88.7,51.3,88.7,56.8z};
    }
}

\newcommand\orcidicon[1]{\href{https://orcid.org/#1}{\mbox{\scalerel*{
                \begin{tikzpicture}[yscale=-1,transform shape]
                \pic{orcidlogo};
                \end{tikzpicture}
            }{|}}}}

\jairheading{}{}{}{}{}
\ShortHeadings{DOCTOR}
{Li \& Jha}
\begin{document}

\title{DOCTOR: A Multi-Disease Detection Continual Learning Framework Based on Wearable Medical Sensors} 
\author{\name Chia-Hao Li $^{\textsuperscript{\orcidicon{0000-0001-9557-6050}}}$ \email chli@princeton.edu \\
        \addr Dept. of Electrical \& Computer Engineering, Princeton University\\
        Princeton, NJ 08544 USA\\
        \name Niraj K. Jha $^{\textsuperscript{\orcidicon{0000-0002-1539-0369}}}$, \textit{Fellow, IEEE} \email jha@princeton.edu \\
        \addr Dept. of Electrical \& Computer Engineering, Princeton University\\
        Princeton, NJ 08544 USA}
\maketitle

\begin{abstract}
Modern advances in machine learning (ML) and wearable medical sensors (WMSs) in edge devices have enabled ML-driven disease
detection for smart healthcare. Conventional ML-driven methods for disease detection rely on customizing individual models
for each disease and its corresponding WMS data. However, such methods lack adaptability to distribution shifts and new
task classification classes. In addition, they need to be rearchitected and retrained from scratch for each new disease.
Moreover, installing multiple ML models in an edge device consumes excessive memory, drains the battery faster, and
complicates the detection process. To address these challenges, we propose DOCTOR, a multi-disease detection continual
learning (CL) framework based on WMSs. It employs a multi-headed deep neural network (DNN) and a replay-style CL algorithm.
The CL algorithm enables the framework to continually learn new missions where different data distributions, classification
classes, and disease detection tasks are introduced sequentially. It counteracts catastrophic forgetting with either a data 
preservation (DP) method or a synthetic data generation (SDG) module. The DP method preserves the most informative subset
of real training data from previous missions for exemplar replay. The SDG module models the probability distribution of the
real training data and generates synthetic data for generative replay while retaining data privacy. The multi-headed DNN
enables DOCTOR to detect multiple diseases simultaneously based on user WMS data. We demonstrate DOCTOR's efficacy in
maintaining high disease classification accuracy with a single DNN model in various CL experiments. In complex scenarios,
DOCTOR achieves 1.43$\times$ better average test accuracy, 1.25$\times$ better F1-score, and 0.41 higher backward transfer
than the naive fine-tuning framework, with a small model size of less than 350KB. \footnote[2]{\scriptsize This work has been submitted to the ACM for possible publication. Copyright may be transferred without notice, after which this version may no longer be accessible.\hfill} \footnote[3]{\scriptsize \copyright~2023 ACM. Personal use of this material is permitted. Permission from ACM must be obtained for all other uses, in any current or future media, including reprinting/republishing this material for advertising or promotional purposes, creating new collective works, for resale or redistribution to servers or lists, or reuse of any copyrighted component of this work in other works.\hfill}
\end{abstract}

% \begin{IEEEImpStatement}
% While conventional ML-driven disease detection methods can achieve good performance in disease detection, installing multiple ML models in an edge device consumes excessive memory, drains the battery faster, and complicates the detection process. The DOCTOR framework addresses these challenges. DOCTOR can continually learn to detect various diseases with a single model and perform simultaneous multi-disease detection by employing the proposed replay-style CL algorithm and multi-headed architecture. DOCTOR achieves a very competitive performance (a 91.6\% average test accuracy and a 96.1\% average macro F1-score) compared to the ideal joint-training framework (a 92.2\% average test accuracy and a 96.9\% average macro F1-score) after consecutively learning three distinct disease detection tasks while maintaining a small model size of less than 350KB. DOCTOR offers a promising approach to smart healthcare by enabling efficient and effective out-of-clinic disease detection. 
% \end{IEEEImpStatement}

% \begin{IEEEkeywords}
% Catastrophic Forgetting, Continual Learning, Disease Detection, Machine Learning, Smart Healthcare, Wearable Medical Sensors.
% \end{IEEEkeywords}

\section{Introduction}
\label{sec:intro}
Physical illnesses and mental health problems impact the well-being of billions of people around the
globe. For centuries, disease detection has been a stressful and time-consuming process for patients.
Patients have to visit a physician and undergo a series of physical and even invasive examinations.
Fortunately, the emergence of wearable medical sensors (WMSs) and modern advances in artificial
intelligence (AI) and machine learning (ML) point to a promising approach to address these problems.
WMSs enable continuous monitoring of physiological signals in a passive, user-transparent, and
non-invasive manner. A sophisticated ML model can then analyze the data captured by WMSs and perform
efficient and effective disease detection. This enables disease detection in an out-of-clinic setting
\cite{covidD, diabD, mhD, cardio, liver, skinD, breast}. 

In real-world scenarios, health record datasets are governed by numerous laws and
restrictions to ensure patient data privacy. Hence, smart healthcare application designers are often
granted access to these datasets for only a limited period of time with restricted or no permission to
preserve data from them. Conventional ML-driven disease detection methods rely on customizing an ML model
for each disease based on its associated WMS dataset. Once new datasets collected from different domains,
containing new classification classes, or gathered for different disease detection tasks become available,
conventional methods require access to both new and old datasets to update the models. Simply retraining
the models with new datasets with conventional methods causes the models to overfit on new data and suffer
from performance deterioration for previously learned missions. This phenomenon is well known as
\textbf{\textit{catastrophic forgetting}} \cite{cataforget}. Therefore, losing access to previously used
datasets hinders conventional methods from adapting to changes in data distributions and additions of new
classification classes for a given disease. In addition, it reduces the flexibility of a trained model 
for learning to detect a distinct disease. For instance, an ML model trained with a dataset collected from 
the elderly might not accurately detect the disease in young adults. A model trained with data gathered from 
healthy individuals and symptomatic COVID-19 patients might be unable to detect the virus in asymptomatic 
patients. These models suffer from performance deterioration on previous datasets when they are retained 
using conventional methods on new datasets after losing access to prior ones. In addition, with conventional 
methods, a model trained for detecting one disease cannot be updated to detect other diseases while maintaining 
performance for the original disease after losing access to its dataset. Therefore, new models need to be 
designed and trained from scratch to detect new diseases. However, placing multiple disease-detecting models 
on edge devices (where WMSs reside) significantly increases memory footprint and drains the battery much faster. 
Consequently, it makes conventional ML-driven disease detection methods suboptimal.

A promising solution to the problems mentioned above is to employ a new framework based on continual learning (CL). The aim
of CL is to alleviate the impact of catastrophic forgetting by preventing
models from overfitting on new data and forgetting the learned knowledge from previous missions \cite{review, overview, survey_a, survey_b, survey_c, survey_d}. CL allows models to adapt incrementally to new data distributions, task classes, and tasks. Thus, CL methods have begun attracting increasing attention. These methods include imposing restrictions on the update process of model parameters through regularization \cite{ewc, si, lwf}, dynamically reconfiguring the model architecture to accommodate new missions \cite{packnet, den, spacenet}, and replaying exemplars from past missions during training \cite{icarl, gem, remind, dgr}. However, most existing approaches focus on image-based CL missions instead of sequential missions based on tabular data \cite{review, overview, survey_a, survey_b, survey_c, survey_d, l2p, CLwVis}, where the data modality is significantly different. Hence, the application of CL methods to tabular datasets is still in its infancy \cite{tabulardata, CL4Recurrent}. 

\begin{figure}[t]
    \centering
    \begin{subfigure}[b]{0.44\textwidth}
        \centering
        \includegraphics[width=\textwidth]{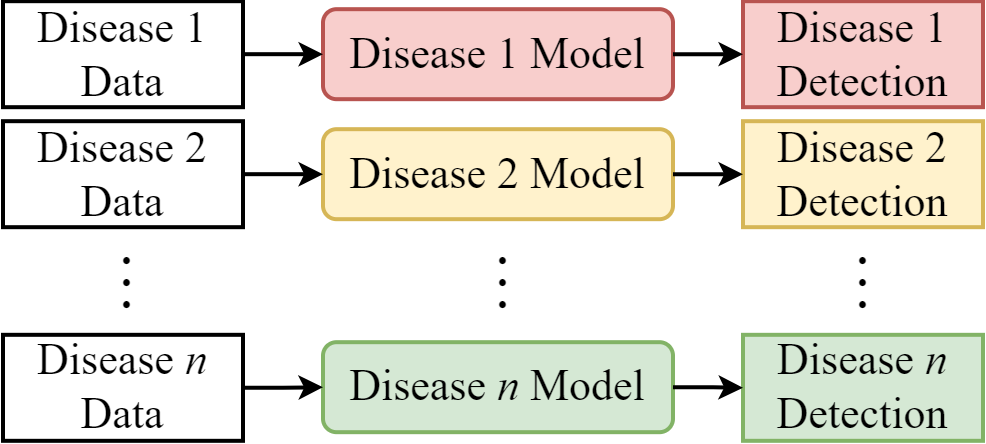}
        \caption{Conventional Disease Detection Methods}
    \end{subfigure}
    \begin{subfigure}[b]{0.55\textwidth}
        \centering
        \includegraphics[width=\textwidth]{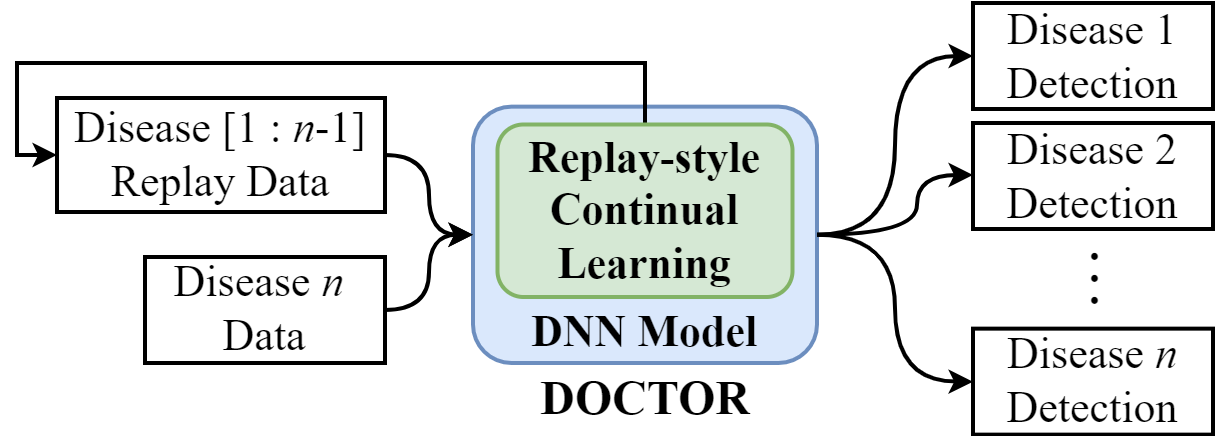}
        \caption{DOCTOR Framework}
    \end{subfigure}
    \caption{An overview of frameworks for disease detection: (a) conventional disease detection methods, and (b) DOCTOR framework. DOCTOR learns to detect new diseases continually and performs multi-disease detection with a single DNN model, whereas conventional methods require a customized model for each disease.}
    \label{fig:overview}
\end{figure}

To address the aforementioned challenges, we propose DOCTOR, a multi-disease detection CL framework based on WMSs. Fig.~\ref{fig:overview} gives an overview of our framework and compares it to conventional ML-driven disease detection methods. DOCTOR performs multi-disease detection with a deep neural network (DNN) and a replay-style CL algorithm. The CL algorithm mitigates catastrophic forgetting in the tabular data domain by training the DNN in an exemplar- or a generative-replay fashion. It enables the framework to continually learn new missions with a single DNN where different data distributions, classification classes, and disease detection tasks are introduced sequentially. Our framework is applicable to other generic DNN models, including recurrent neural networks (RNNs) \cite{RNN} and long short-term memory (LSTM) networks \cite{LSTM}. In addition, DOCTOR does not require manual feature extraction procedures. It operates directly on sequential time series tabular data collected from commercially available WMSs.

The proposed replay-style CL algorithm employs either a data preservation (DP) method or a synthetic data generation (SDG) module for a given
disease. The DP method preserves the most informative subset of real training data from previous missions for exemplar replays based on their average training loss. It enables efficient sampling without incurring excessive computational costs when preserving real data is allowed. The SDG module recreates the joint multivariate probability distributions of the real training data from previous missions using both parametric and non-parametric density estimation methods. Subsequently, the module generates synthetic data from the learned probability distributions of previous missions for generative replays during training. Therefore, the SDG module can generate as much training data as needed when preserving real data is infeasible due to memory restrictions or data privacy issues. Moreover, it reduces memory footprint by storing only the learned probability distribution rather than the generated synthetic data. 

We adopt a multi-headed DNN architecture in our framework \cite{si, lwf, CLwVis}. When DOCTOR learns a
new disease detection task, it first creates a new detection head parallel to those trained for
previous tasks. Then, it learns an individual classification probability distribution for the new task
at the new head. Meanwhile, the other heads for the previous tasks are fine-tuned through replays. Due
to the individual probability distribution learned at each detection head, the multi-headed
architecture allows DOCTOR to detect multiple diseases simultaneously based on user WMS data. We
demonstrate its efficacy with three different disease datasets under three distinct CL scenarios.
To the best of our knowledge, this is the first CL framework for multi-disease detection 
with WMS data. Novelty also lies in our efficient sampling method for exemplar replay and 
application of density-estimation-based SDG for generative replay. 

The rest of the article is organized as follows. Section \ref{sec:back_rel} discusses the background and related works on ML-driven disease detection, CL algorithms, and probability density estimation. Section \ref{sec:def_metrics} elaborates on the problem definitions and evaluation metrics used in our work. Section \ref{sec:doctor} gives details of our proposed framework. Section \ref{sec:setup} provides details of the disease datasets used in our experiments and framework implementation. Section \ref{sec:results} presents the experimental results. Section \ref{sec:dis_lim} discusses the limitations of our framework and possible solutions. Finally, Section \ref{sec:conclusion} concludes the article.

\section{Background}
\label{sec:back_rel}
In this section, we introduce related works in the fields of ML-driven disease detection 
and CL algorithms. In addition, we provide background on probability density estimation.

\subsection{Machine Learning for Disease Detection}
Modern developments in ML enable efficient and effective ML-driven disease detection systems. For example, CovidDeep \cite{covidD} and MHDeep \cite{mhD} detect the SARS-CoV-2 virus/COVID-19 disease and mental health disorders based on physiological signals collected from commercially available WMSs and smartphones. They also employ a grow-and-prune algorithm \cite{SCANN} to synthesize optimal DNN architectures for detection. Similarly, DiabDeep \cite{diabD} uses grow-and-prune synthesis to deliver DNN models with sparsely connected layers or sparsely recurrent layers for diabetes detection based on WMS data and patient demographic information. Miao and Miao \cite{cardio} present a DNN model that performs fetal health assessment in complications of pregnancy based on multi-class morphological pattern predictions with cardiotocography data. Ghazal \textit{et al.} \cite{liver} train a DNN to predict early-stage liver disease with tabular patient data. Dritsas and Trigka use a rotation forest model to predict chronic kidney disease with patient physiological data. Allugunti \cite{skinD} employs a convolutional neural network (CNN) to classify different types of melanoma skin diseases using patient skin images. Finally, Melekoodappattu \textit{et al.} \cite{breast} combine a customized CNN with image texture attribute extraction to perform breast cancer detection in mammograms. 

\subsection{Continual Learning Algorithms}
Many CL algorithms have been proposed in the literature to counteract the catastrophic forgetting phenomenon in ML models. Recent CL algorithms can mainly be categorized into three groups: regularization-based, architecture-based, and replay-based \cite{review, overview, survey_a, survey_b, survey_c, survey_d}. 

\subsubsection{Regularization-based Methods} 
These methods use data-driven regularization to impose restrictions on the update process of model parameters to prevent performance deterioration on the learned missions. One canonical example of these methods is \emph{elastic weight consolidation} (EWC) \cite{ewc}. EWC adopts a quadratic penalty on parameters that are important to previously learned missions for deviating from the learned values. The importance of the learned parameters is weighted by the diagonal of each Fisher information matrix of each learned mission.

Another example of such a method is \emph{synaptic intelligence} (SI) \cite{si}. In this work, each intelligent synapse estimates its own importance in solving a given mission based on its update trajectory. Then, SI uses this information as the weighting parameter for its quadratic regularization function to decelerate the update process to mitigate catastrophic forgetting for previously learned missions.

Apart from restricting the model parameter update process, \emph{learning without forgetting} (LwF) \cite{lwf} draws inspiration from knowledge distillation in a multi-headed architecture setting. Before learning a new mission, LwF records the current network's output response to new data to generate pseudo labels. Subsequently, the pseudo labels are utilized in the training process for regularization to distill knowledge learned from previous missions.

Regularization-based methods address catastrophic forgetting without storing exemplars and
incurring additional memory overhead. However, they do not achieve satisfactory performance under challenging CL
settings \cite{CLwVis} or complex datasets \cite{largeIL}. Thus, we do not use regularization-based CL algorithms in our framework.

\subsubsection{Architecture-based Methods} 
These methods modify the model architecture in various ways to accommodate new missions while maintaining learned knowledge 
from previous ones. For instance, PackNet \cite{packnet} interactively prunes the unimportant parameters after learning each mission and then retrains the network to spare some connections for future missions. In addition, a parameter selection mask is stored for each learned mission to specify the connections reserved for the mission at test time.

\emph{Dynamically expandable network} (DEN) \cite{den} dynamically expands network
capacity to compose a compact, overlapping, and knowledge-sharing structure to learn new missions. DEN exploits
a multi-stage process, which first selects the relevant parameters from previous missions to optimize and then evaluates the loss on the new mission after training. Once the loss exceeds a designated threshold, DEN expands network 
capacity and trains the expanded network with group sparsity regularization to avoid excessive growth.

On the other hand, SpaceNet \cite{spacenet} exploits the fixed capacity of a sparse neural network. 
It effectively frees up space in the model for future missions and produces sparse representations in the hidden 
layers for each learned mission to reduce inter-mission interference. When learning a new mission, it employs an adaptive training method to train the network from scratch, where neurons important to learned missions are reserved while others are shared in learning future missions.

Architecture-based methods perform CL by expanding or sparing model capacity to learn new missions while memorizing
previous ones. Nevertheless, they might require a substantial amount of additional parameters, be more complex to train,
and not scale to a large number of missions. Hence, they are generally not feasible for wearable edge devices. As a result,
we do not select architecture-based CL methods for our framework.

\subsubsection{Replay-based Methods} 
Generally, these methods maintain a small buffer to store sampled exemplars from previous
missions or exploit generative models to generate synthetic samples representing learned missions. These samples
are then replayed when learning a new mission to retain the knowledge learned in previous missions and
alleviate catastrophic forgetting. One of the first examples is \emph{incremental classifier and representation
learning} (iCaRL) \cite{icarl}. For each class, iCaRL selects a subset of exemplars that best approximates their
class center in the latent feature space and stores them in a memory buffer. Then, the stored samples are
replayed with new data to perform nearest-mean-of-exemplars classification when learning a new class and at test
time. In addition, the distance between data instances in the latent feature space is used to update the memory buffer. 
However, despite how simple and effective the idea is, iCaRL focuses on the class-incremental CL scenario only and is not 
applicable to challenging CL settings, such as the task-incremental scenario (see details in Section \ref{sec:def}).

\emph{Gradient episodic memory} (GEM) \cite{gem} is a method that stores exemplars from past
missions in the episodic memory buffer and exploits them to compute past mission gradients. When learning a new
mission, GEM solves a constrained optimization problem that projects the current mission gradients in a feasible
area outlined by the past mission gradients. It alleviates catastrophic forgetting by preventing the model
parameters from updating in a direction that increases the loss of each previous mission. GEM exploits exemplar
replay and combines it with the regularizing parameter update process, but it does not perform well on complex 
datasets and in challenging CL settings \cite{survey_b, survey_d}.

Apart from storing raw exemplars, \emph{replay using memory indexing} (REMIND) \cite{remind} draws inspiration from
hippocampal indexing theory to store compressed data samples. By performing tensor quantization, REMIND efficiently stores
compressed representations of data instances from previous missions instead of raw images for future replays. It not only
increases buffer memory efficiency but also improves the scalability of the framework. Nevertheless, 
this work targets the image classification domain and cannot be applied to tabular data directly.

On the other hand, \emph{deep generative replay} \cite{dgr} uses a generative adversarial
network \cite{gan}, a generator, to generate synthetic images that mimic images from previous missions. The
generated images are then paired with their corresponding labels produced by another task-solving model, a
solver, to represent past missions. When learning a new mission, the synthetic images and their corresponding
labels are interleaved with new data to update the generator and solver networks together. Generating synthetic
data as a substitution for exploiting stored real data can address the data privacy issue and give the model access 
to as much training data as needed for exemplar replay. However, this work also focuses on sequential image 
classification missions and cannot be deployed in the tabular data domain.

In general, replay-based methods lead to memory storage overhead to store exemplars. Moreover,
preserving actual data might not always be feasible in real-world applications due to data privacy restrictions
\cite{survey_b, survey_d}. However, empirical evaluations have shown that they achieve the best performance
trade-offs and are much stronger at avoiding catastrophic forgetting than the other two methods, even under
complex CL scenarios \cite{survey_c, tabulardata, perfectmemory}. Therefore, we employ a replay-style CL algorithm 
for sequential time series tabular data in our framework (see details in Section \ref{sec:replayCL}).

\subsection{Probability Density Estimation}
\label{sec:para}
A probability density function (PDF) is a mathematical function that describes the probability 
distribution of a continuous random variable. It specifies the likelihood of the random variable taking on a value 
within a specific interval. A probability density estimation method can be used to estimate the PDF of a continuous
random variable based on observed data. Probability density estimation methods can be categorized into two groups: 
parametric and non-parametric.

\subsubsection{Parametric Density Estimation}
In parametric density estimation, a PDF is assumed to be a member of a parametric family. Hence, density estimation is transformed into finding estimates of the various parameters of the parametric family. The canonical example of a parametric density estimation method is the Gaussian mixture model (GMM). Let $x_1, x_2, \ldots, x_N$ be independent and identically distributed (\emph{i.i.d.}) samples drawn from some univariate distribution with an unknown density $f$ at any given point $x$. The approximated PDF of $f$ is modeled as a mixture of $C$ Gaussian models in the form of:

\begin{equation*}
    p_X(x|\Theta) = \sum_{c=1}^Cp_{X|Z}(x|z_c,\Theta)\,p_Z(z_c|\Theta),
\end{equation*}

\noindent where $X$ is the observed variables, $\Theta$ represents the parameters of the GMM, $c$ is the
Gaussian model component, $z_c$ denotes the hidden state variable of $c$, and $Z$ symbolizes the hidden state variables that indicate the GMM assignment. The prior probability of each Gaussian model component $c$ can be written as:

\begin{equation*}
    p_Z(z_c|\Theta) = \theta_c.
\end{equation*}

\noindent Each Gaussian model component $c$ is a $d$-dimensional multivariate Gaussian distribution with mean vector $\mu_c$ and covariance matrix $\Sigma_c$ in the form of:

\begin{equation*}
\begin{split}
    p_{X|Z}(x|z_c,\Theta) = &\left(\frac{1}{2\pi}\right)^{d/2}|\Sigma_c|^{-1/2}\exp\left(-\frac{1}{2}(x-\mu_c)^T\,\Sigma_c^{-1}\,(x-\mu_c)\right),
\end{split}
\end{equation*}

\noindent where $|\Sigma_c|$ is the determinant of the covariance matrix. Therefore, the PDF modeled by GMM can be simplified as follows:

\begin{equation}
\label{eq:1}
    p_{X|C}(x) = \sum_{c=1}^C\pi_c\mathcal{N}(x|\mu_c,\Sigma_c),
\end{equation}

\noindent where $\pi_c$ represents the weight of the Gaussian model component $c$ and $\mathcal{N}$ denotes the normal distribution.

\subsubsection{Non-parametric Density Estimation}
\label{sec:non-para}
While a non-parametric density estimation scheme does not make any distributional assumptions, it can generate
any possible PDF from observed data. It covers a broad class of functions that can approximate the probability 
distribution of a continuous random variable. One classic example is kernel density estimation (KDE). KDE approximates a probability distribution as a sum of many kernel functions. Let $x_1, x_2, \ldots, x_N$ be \emph{i.i.d.} samples drawn from some univariate distribution with an unknown density $f$ at any given point $x$. The approximated PDF $\hat{f}$ with the KDE method can be formulated as follows:

\begin{equation*}
    \hat{f}(x) = \frac{1}{Nh}\sum_{i=1}^NK(\frac{x-x_i}{h}),
\end{equation*}

\noindent where $N$ is the total number of samples, $x_i$ is the $i$-th sample, $K$ represents the underlying kernel function, and $h$ symbolizes the kernel bandwidth. There may be various options for choosing the kernel function $K$, for instance, the PDF of the normal distribution. However, each kernel function should satisfy the following property:

\begin{equation*}
    \int_{-\infty}^{\infty} K(x)\,dx = 1, \; K(x) > 0 \;\; \forall x.
\end{equation*}

\noindent Another important design parameter for KDE is the kernel bandwidth $h$. It scales $K$ and controls the smoothness of the estimated function. In general, $h$ has a strong impact on the bias-variance trade-off of the KDE. Essentially, high-variance models estimate the training data well but suffer from overfitting on the noisy training data. On the other hand, high-bias models are simpler but suffer from underfitting on the training data. Whereas $h$ has to be 
close to 0 to achieve a small bias, it needs to be close to $\infty$ to achieve a small variance.

\section{Problem Definitions and Evaluation Metrics}
\label{sec:def_metrics}
In this section, we first formalize the problem definitions for CL tasks. Then, we introduce the evaluation metrics 
used to evaluate the performance of our framework. 

\subsection{Problem Definitions}
\label{sec:def}
CL is usually defined as training a single ML model on a sequence of missions where non-stationary data distributions,
different classification classes, or distinct tasks are presented sequentially \cite{overview, l2p}. In addition, data from
previous missions may no longer be available while training on current and future missions. We define a potentially infinite
sequence of missions as $M = \{M_1, M_2, \ldots, M_n, \ldots\}$, where the $n$-th mission is depicted by $M_n = \{(X_n^i, Y_n^i)\ |\
i = 1, 2, \ldots, C_n\}$. $X_n^i \in \mathcal{X}$ and $Y_n^i \in \mathcal{Y}$ refer to the set of features and their corresponding
class labels for the $i$-th class in mission $M_n$. $C_n$ refers to the total number of classes in mission $M_n$. The objective
of CL is to train a single model $f_\theta: \mathcal{X} \rightarrow \mathcal{Y}$, parameterized by $\theta$, such that it can
sequentially learn new missions in $M$ and predict their labels $Y = f_\theta(X \in \mathcal{X}) \in \mathcal{Y}$ without
degrading the performance of previous missions.

Depending on the mission transition scenario, CL can usually be categorized into three different settings: (i) domain-, (ii) class-, and (iii) task-incremental CL \cite{scenario}. 

\textbf{Domain-incremental CL} represents the scenario where data from different distribution domains for the same task become available sequentially. In other words, mission $M_n$ contains the same task and classification classes as missions $[M_1:M_{n-1}]$, but $X_n^i$ and $[X_1^i:X_{n-1}^i]$ come from different probability distributions. For example, data collected from different cities, countries, or age groups become available incrementally for the same disease. 

\textbf{Class-incremental CL} describes the setting where new classification classes of the current task emerge
incrementally. To put it another way, mission $M_n$ encompasses the same task as missions $[M_1:M_{n-1}]$, yet $Y_n^i$ includes classes unseen in $[Y_1^i:Y_{n-1}^i]$. For instance, some patients are diagnosed with COVID-19 while not experiencing any symptoms and are later categorized as asymptomatic-positive. 

\textbf{Task-incremental CL} refers to the case where new classification tasks appear in a sequence. That is, mission $M_n$ contains new classification tasks that are unseen in missions $[M_1:M_{n-1}]$. For example, mission $M_n$ may contain the COVID-19 detection task, whereas missions $[M_1:M_{n-1}]$ may comprise other disease detection tasks. 

To evaluate the efficacy of DOCTOR, we conducted extensive experiments in all three settings with three different disease
datasets obtained using commercially available WMSs. Section \ref{sec:results} gives more details about how these three settings are used in our work.

\subsection{Evaluation Metrics}
\label{sec:metrics}
Several metrics are popularly used to evaluate CL algorithms in the literature, such as average accuracy, average forgetting, backward transfer (BWT), and forward transfer \cite{overview, survey_d, gem}. To illustrate DOCTOR's overall performance in continually learning multiple disease detection missions, we report the average accuracy and average F1-score across all learned missions in our experimental results. When reporting the average F1-score, we define true positives (negatives) as the unhealthy (healthy) instances correctly classified as disease-positive (healthy) and false positives (negatives) as the healthy (unhealthy) instances misclassified as disease-positive (healthy).

In addition, to evaluate the proposed replay-style CL algorithm for mitigating catastrophic forgetting, we also report
the BWT value in our experimental results. When learning a new mission, BWT measures how much the CL algorithm impacts the 
performance of the framework on previous missions. BWT can be defined as:

\begin{equation*}
    \text{BWT} = \frac{1}{q-1}\sum_{n=1}^{q-1}(a_n^q - a_n^n),
\end{equation*}

\noindent where $a_n^q$ represents the test accuracy of the framework for the $n$-th mission after continually learning a 
total of $q$ missions, and $a_n^n$ denotes the test accuracy of the framework for the $n$-th mission after continually 
learning a total of $n$ missions.

\begin{figure}[t]
    \centering
    \begin{subfigure}[b]{0.51\textwidth}
        \centering
        \includegraphics[width=\textwidth]{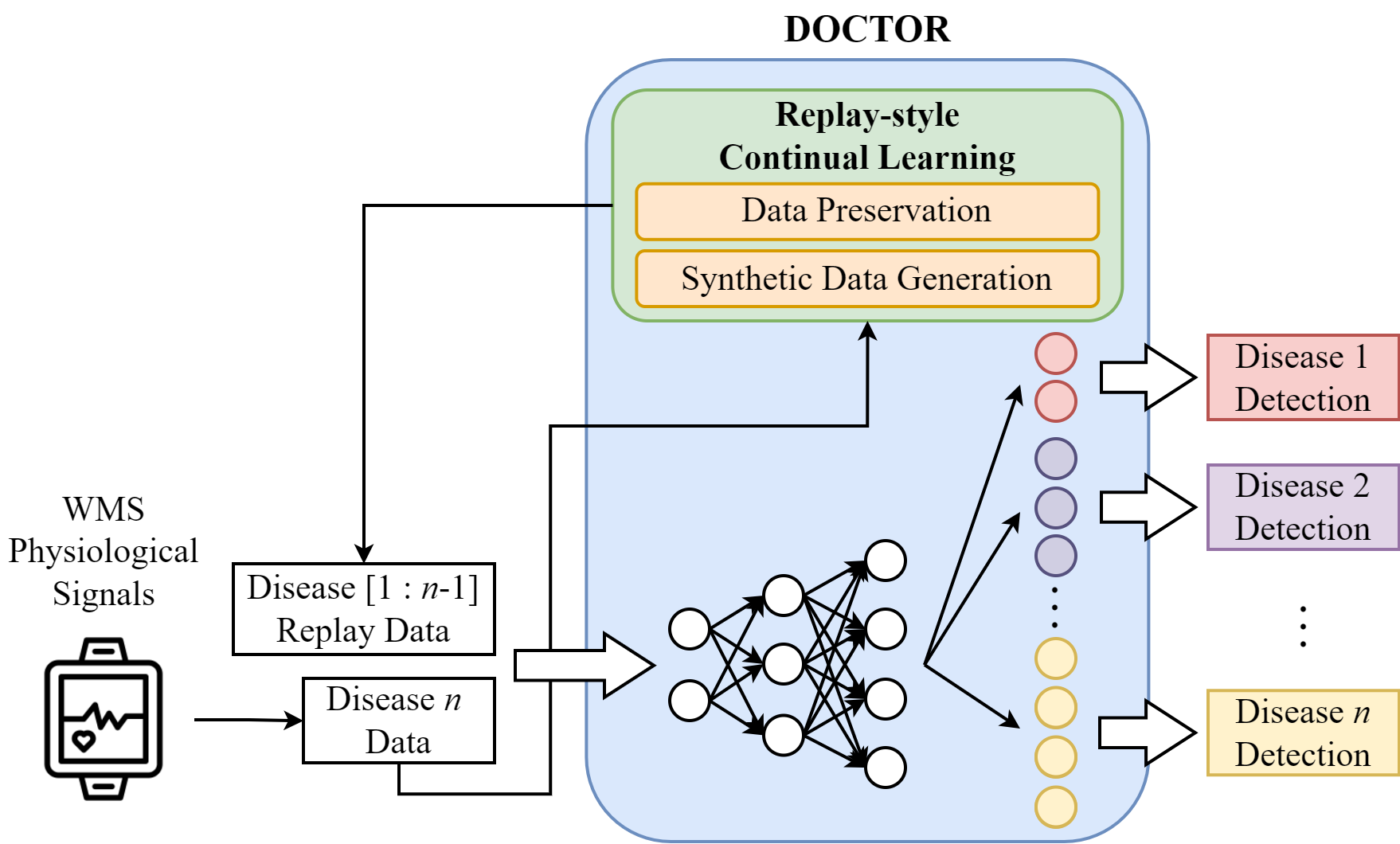}
        \caption{Schematic Diagram}
        \label{fig:schematic}
    \end{subfigure}
    \begin{subfigure}[b]{0.48\textwidth}
        \centering
        \includegraphics[width=\textwidth]{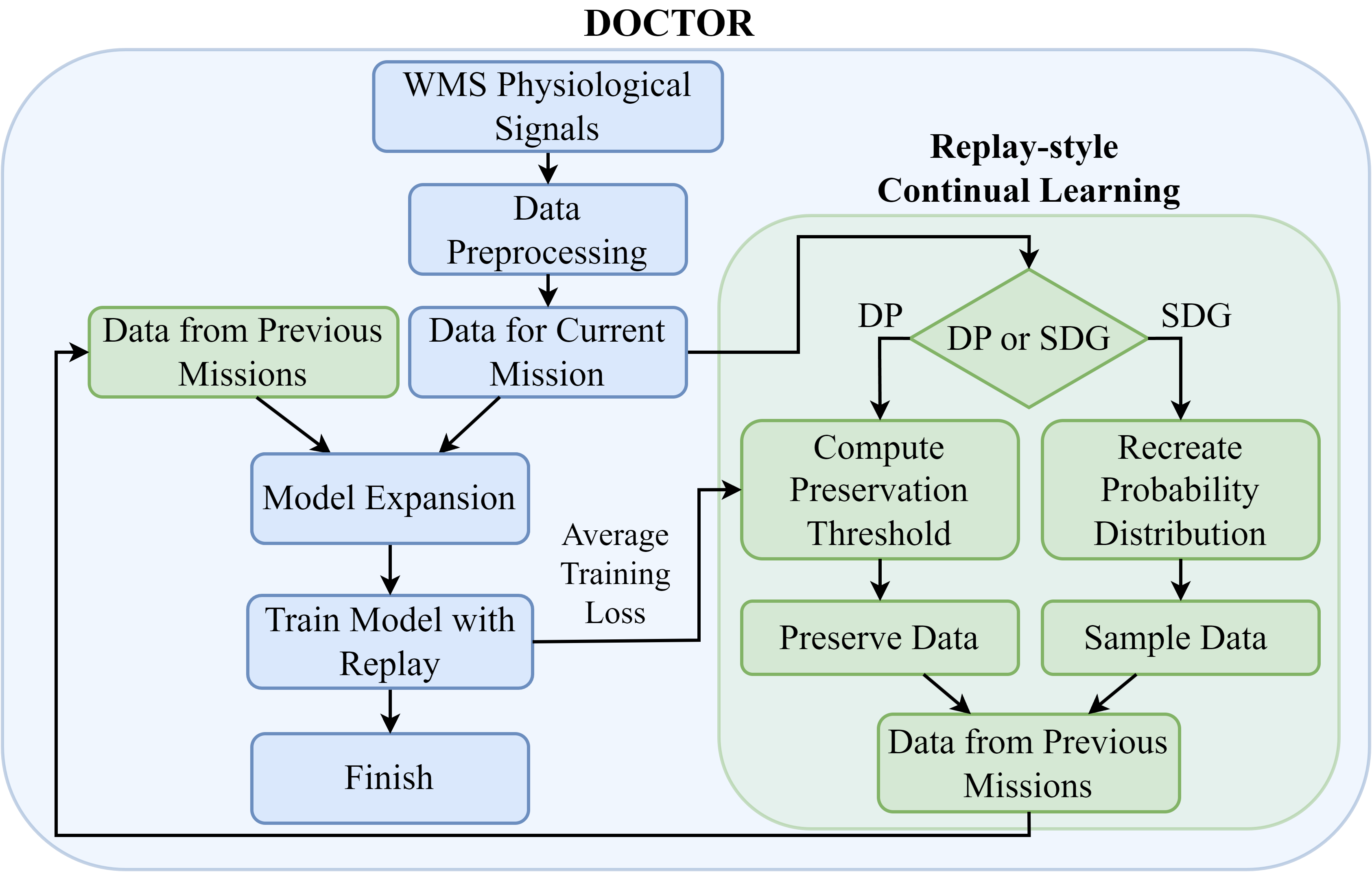}
        \caption{Top-level Flowchart}
        \label{fig:flowchart}
    \end{subfigure}
    \caption{The Doctor framework: (a) schematic diagram and (b) top-level flowchart. (DP: data preservation, 
SDG: synthetic data generation.)}
    \label{fig:diagram}
\end{figure}

\section{The DOCTOR Framework}
\label{sec:doctor}
In this section, we give details of the DOCTOR framework. We start with an overview and a top-level flowchart. Next, we elaborate on the multi-disease detection workflow of DOCTOR in three different CL scenarios. Finally, we provide details of the proposed replay-style CL algorithm. 

\subsection{Framework Overview}
Fig.~\ref{fig:schematic} shows the schematic diagram of the framework. It operates directly on tabular data
collected from commercially available WMSs. It learns multiple disease detection tasks sequentially with a
single multi-headed DNN model through our proposed replay-style CL algorithm. The CL algorithm comprises a DP 
method and an SDG module. When the DNN model learns a new mission, the CL algorithm replays data
from previous missions with the new data to update the model. Finally, the multi-headed architecture generates individual classification probability distribution for each output detection head. It enables DOCTOR to simultaneously detect multiple diseases for patients based on their WMS data. 

Fig.~\ref{fig:flowchart} illustrates the top-level flowchart of DOCTOR operating in a CL mission. First, we collect physiological signals from patients through commercially available WMSs. Then, we apply basic data preprocessing to the raw data, including data stream synchronization, windowing, and standardization, to prepare the datasets. Next, the CL algorithm gathers data from previous missions for replay during training. The data of a previously learned mission can be either real data preserved through the DP method or synthetic data generated by the SDG module. In the model expansion step, the 
framework determines if more output neurons or detection heads are required for the current mission and expands the DNN
model accordingly (details in Section \ref{sec:multi-disease}). Due to differences in training dataset sizes across
missions, the model might be biased toward missions with more available training data. To avoid the biased prediction
problem, we train the model jointly with data from the current mission and previous missions in a balanced fashion. In
other words, we gather an even amount of data from each mission to form a mini-batch for training.
For example, suppose we are learning Mission 3 while employing a set of preserved data from Mission 1 
and a probability density estimation model for Mission 2. For each training iteration with a batch size $B$, the mini-batch 
is composed of $\lfloor \frac{B}{3} \rfloor$ preserved data from Mission 1, $\lfloor \frac{B}{3} \rfloor$ synthetic data 
for Mission 2, and $\lfloor \frac{B}{3} \rfloor$ real training data from Mission 3. In parallel, the user or
practitioner can specify which method to use in the replay-style CL algorithm. If preserving real data is allowed for
the current mission and memory storage is sufficient, the efficient DP method can be used. It obtains the average training loss of each data instance in the current mission after training and computes the threshold value for DP. Then, it preserves data instances whose average training loss values are above the threshold value for future replays (more details in Section \ref{sec:replayCL}). If patient privacy is critical to the current mission, the SDG module can be used to preserve data privacy. It models the joint multivariate probability distributions of the real training data through both parametric and non-parametric density estimation methods. The module stores only the learned probability density estimation models for future replays. Then, the module samples as much synthetic data as required for replay from the learned probability distributions on the fly during training (details in Section \ref{sec:replayCL}).

\begin{figure}[t]
    \centering
    \begin{subfigure}[b]{0.4\textwidth}
        \centering
        \includegraphics[width=\textwidth]{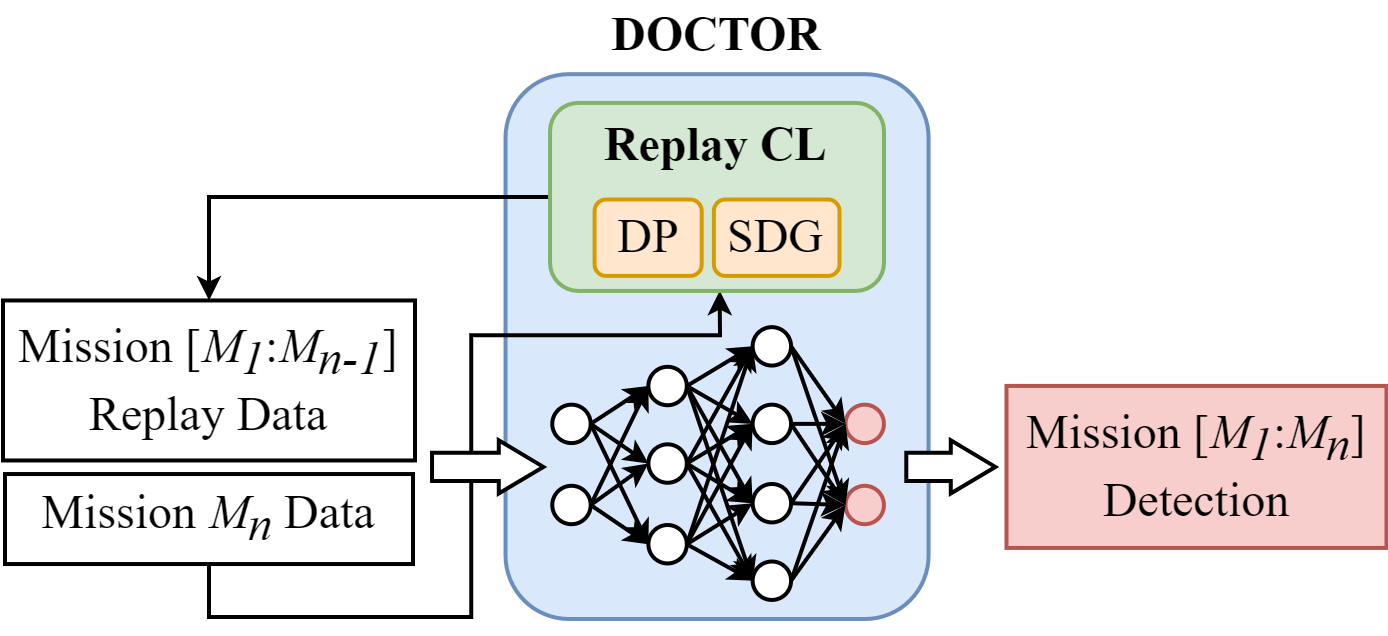}
        \caption{Domain-incremental Scenario}
        \label{fig:DIL}
    \end{subfigure}
    \begin{subfigure}[b]{0.4\textwidth}
        \centering
        \includegraphics[width=\textwidth]{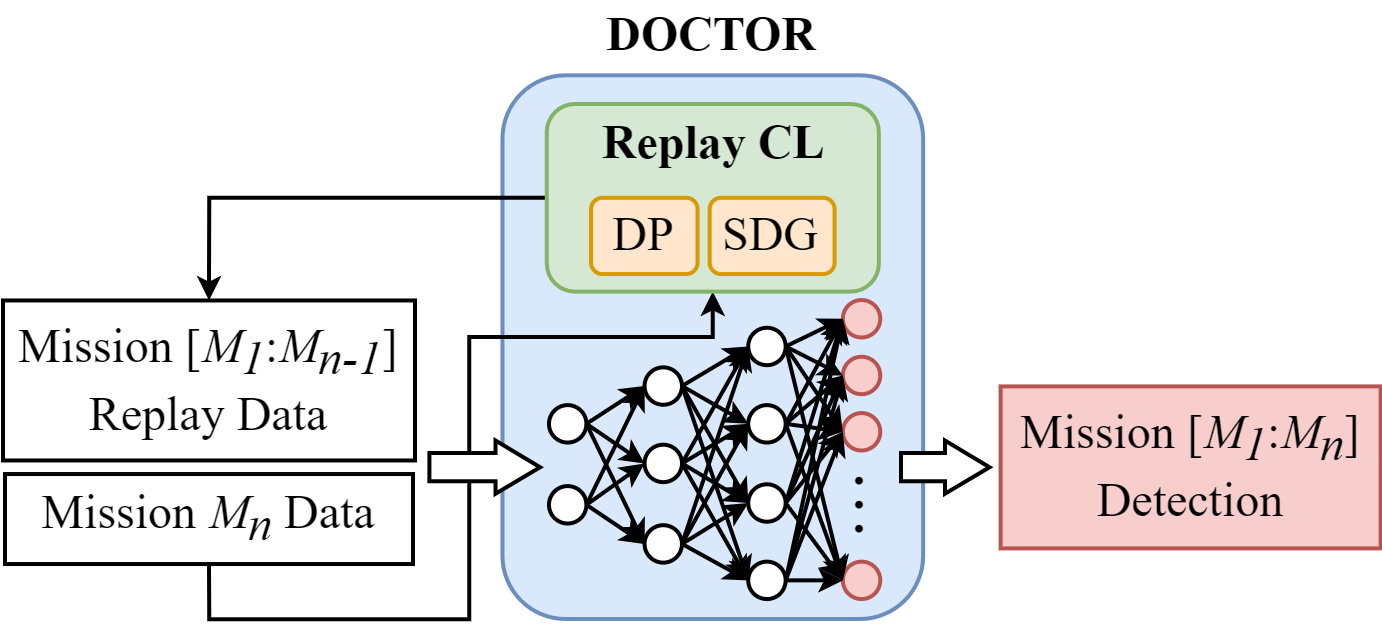}
        \caption{Class-incremental Scenario}
        \label{fig:CIL}
    \end{subfigure}
    \begin{subfigure}[b]{0.4\textwidth}
        \centering
        \includegraphics[width=\textwidth]{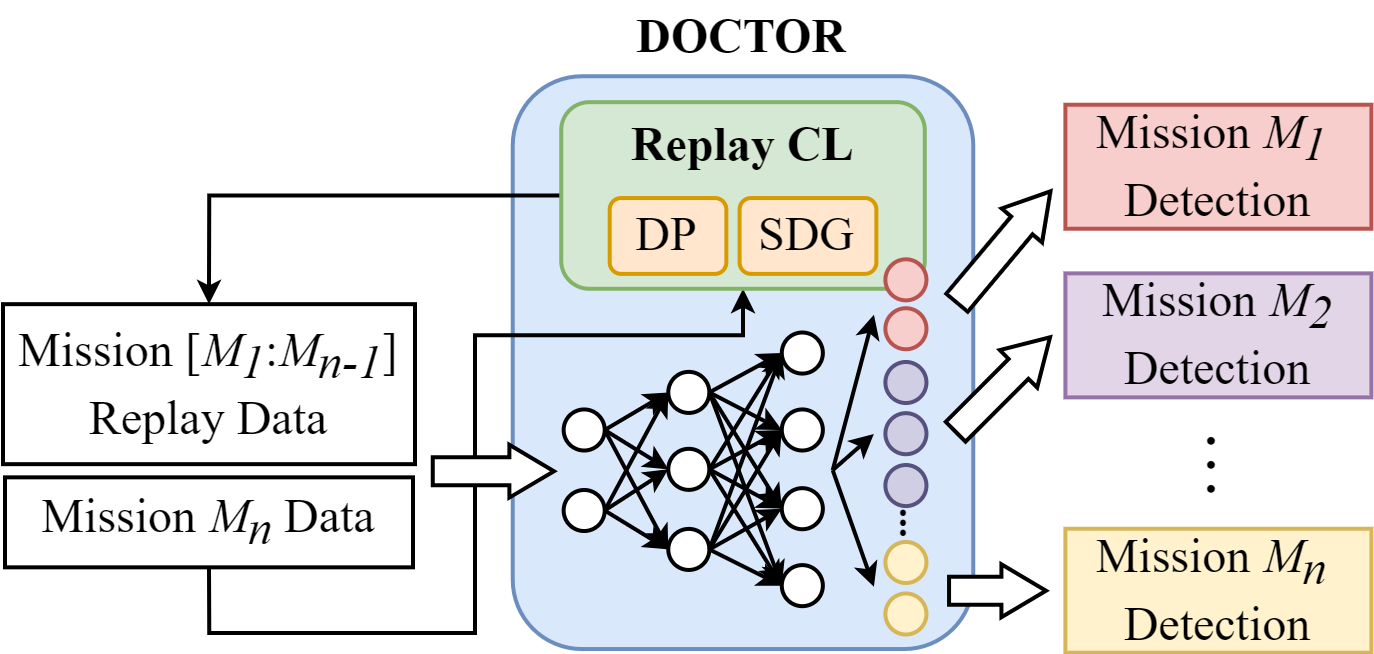}
        \caption{Task-incremental Scenario}
        \label{fig:TIL}
    \end{subfigure}
    \caption{The model expansion procedures for multi-disease detection in the (a) domain-incremental scenario, (b) class-incremental scenario, and (c) task-incremental scenario in DOCTOR. (DP: data preservation, SDG: synthetic data generation.)}
    \label{fig:three_settings}
\end{figure}

\subsection{Multi-disease Detection}
\label{sec:multi-disease}
Next, we elaborate on the multi-disease detection workflows and the DNN model expansion procedures in DOCTOR under three different CL scenarios. We give details of how our framework continually adapts to data distribution domain shifts and new classification classes. We also detail how DOCTOR continually learns to perform new disease detection tasks. Then, we explain its application to multi-disease detection at test time. 

\subsubsection{Domain-incremental Scenario}
Fig.~\ref{fig:DIL} shows the DOCTOR framework after continually learning $n$ missions in the domain-incremental scenario. As described in Section \ref{sec:def}, the new mission $M_n$ contains the same task and classification classes as missions $[M_1:M_{n-1}]$, but $X_n^i$ comes from a different probability distribution than $[X_1^i:X_{n-1}^i]$. Thus, the framework does not need to add new output neurons or create new output detection heads to adapt to new data from different domains. The DNN model simply needs to be fine-tuned jointly with data from the new domain and previous domains to prevent catastrophic forgetting. 

\subsubsection{Class-incremental Scenario}
Fig.~\ref{fig:CIL} illustrates the DOCTOR framework after continually learning $n$ missions in the class-incremental
scenario. In this setting, the new mission $M_n$ includes unseen classification classes in missions $[M_1:M_{n-1}]$ for
the same task, as mentioned in Section \ref{sec:def}. Hence, the framework needs to add new output neurons to the DNN
model to accommodate new classification classes. In addition, the DNN model inherits the trained weights from the last
stage after model expansion. Then, the model gets trained jointly with data from the new mission and previous missions
to learn the new classification classes and retain the knowledge of the previous ones. 

\subsubsection{Task-incremental Scenario}
Fig.~\ref{fig:TIL} demonstrates the DOCTOR framework after continually learning $n$ missions in the task-incremental scenario. As explained in
Section \ref{sec:def}, the new mission $M_n$ includes a different task with completely distinct classification classes from
missions $[M_1:M_{n-1}]$. Hence, DOCTOR needs to incrementally learn to detect various disease detection tasks
in this scenario. To do this, we adopt a multi-headed architecture for the DNN model to accommodate new disease detection tasks. 

As shown in Fig.~\ref{fig:TIL}, when a new disease detection task arrives, the framework generates a new detection head in the output layer of the DNN model in the model expansion step. The new detection head is added in parallel to the other ones. Therefore, it can learn to detect the new disease and perform classification individually without interfering with the other heads. As before, the DNN model inherits the trained weights for the learned tasks from the last stage after expansion. Then, the model is trained with data from the new task and fine-tuned with data from previous tasks through replays at their corresponding detection heads. 

We adopt a softmax layer for each output detection head. Thus, each head can generate an individual probability
distribution for classification and output the predicted detection result for each disease separately. We then obtain
the cross-entropy loss at each head for its corresponding training data instances and their true labels. Finally, we accumulate the losses from all detection heads to perform backpropagation. 

\begin{figure}[t]
    \centering
    \includegraphics[width=0.6\linewidth]{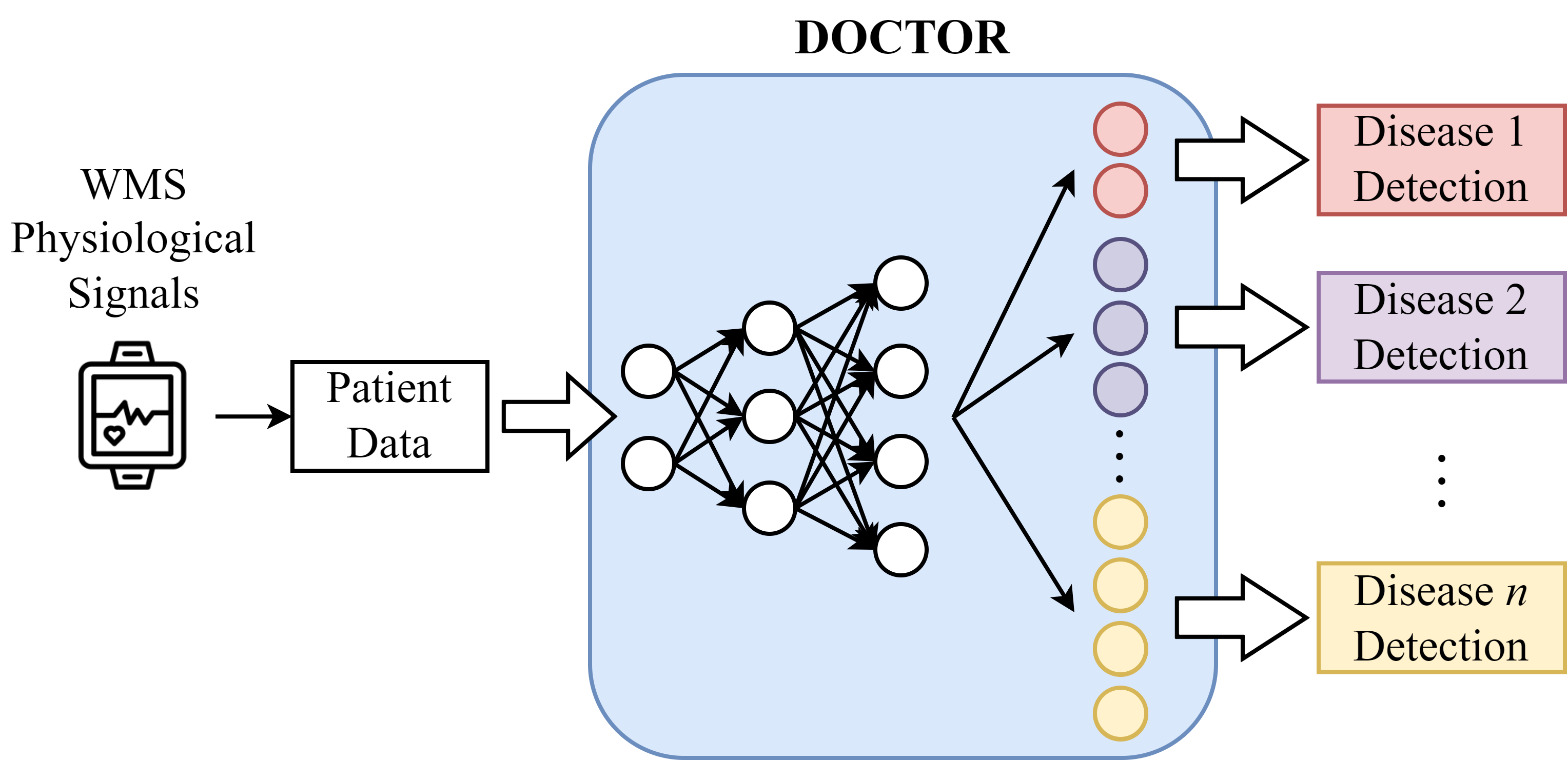}
    \caption{The DOCTOR framework at test time. (DP: data preservation, SDG: synthetic data generation.)}
    \label{fig:test_time}
\end{figure}

\subsubsection{Multi-disease Detection Application}
The multi-headed architecture allows DOCTOR to generate individual classification probability distribution for
each detection head. Therefore, DOCTOR can perform detection for different diseases in parallel.
Fig.~\ref{fig:test_time} demonstrates how DOCTOR can be used for multi-disease detection. At test time,
physiological signals are collected with commercially available WMSs and sent to the framework. The framework
then preprocesses the raw data and performs disease detection with the multi-headed DNN model. Finally, DOCTOR outputs the detection result for each disease from each detection head. Therefore, patients can be informed about which diseases are detected from their physiological signals with a single examination. For example, a patient may learn that he or she is asymptomatic-positive for COVID-19, Type-II diabetic, and mentally depressed through a single test. 

\subsection{Replay-style Continual Learning Algorithm}
\label{sec:replayCL}
The CL algorithm we propose for our DOCTOR framework targets the tabular data domain and
counteracts catastrophic forgetting with replays. It consists of a DP method and an SDG module, which 
can be deployed based on the use scenario and user settings. The DP method preserves the most 
informative subset of real training data from previous missions for exemplar replay. The SDG module can generate synthetic 
data from the learned multivariate probability distributions of past missions for generative replay. Next, we give 
details of the DP method and SDG module employed in our replay-style CL algorithm.

\subsubsection{Data Preservation}
\label{sec:DP}
In scenarios where computational resources are limited and preserving actual data is allowed, DOCTOR 
can employ the DP method for CL. The method preserves the most informative subset of real training data 
and their labels from previous missions in a buffer for future replays. It samples from real training data in a 
stratified fashion based on the average training loss value of each training data instance. The preserved 
data are replayed together with data from the new mission in future CL scenarios to retain the knowledge extracted from 
previous missions.

\begin{figure}[t]
    \centering
    \includegraphics[width=0.7\linewidth]{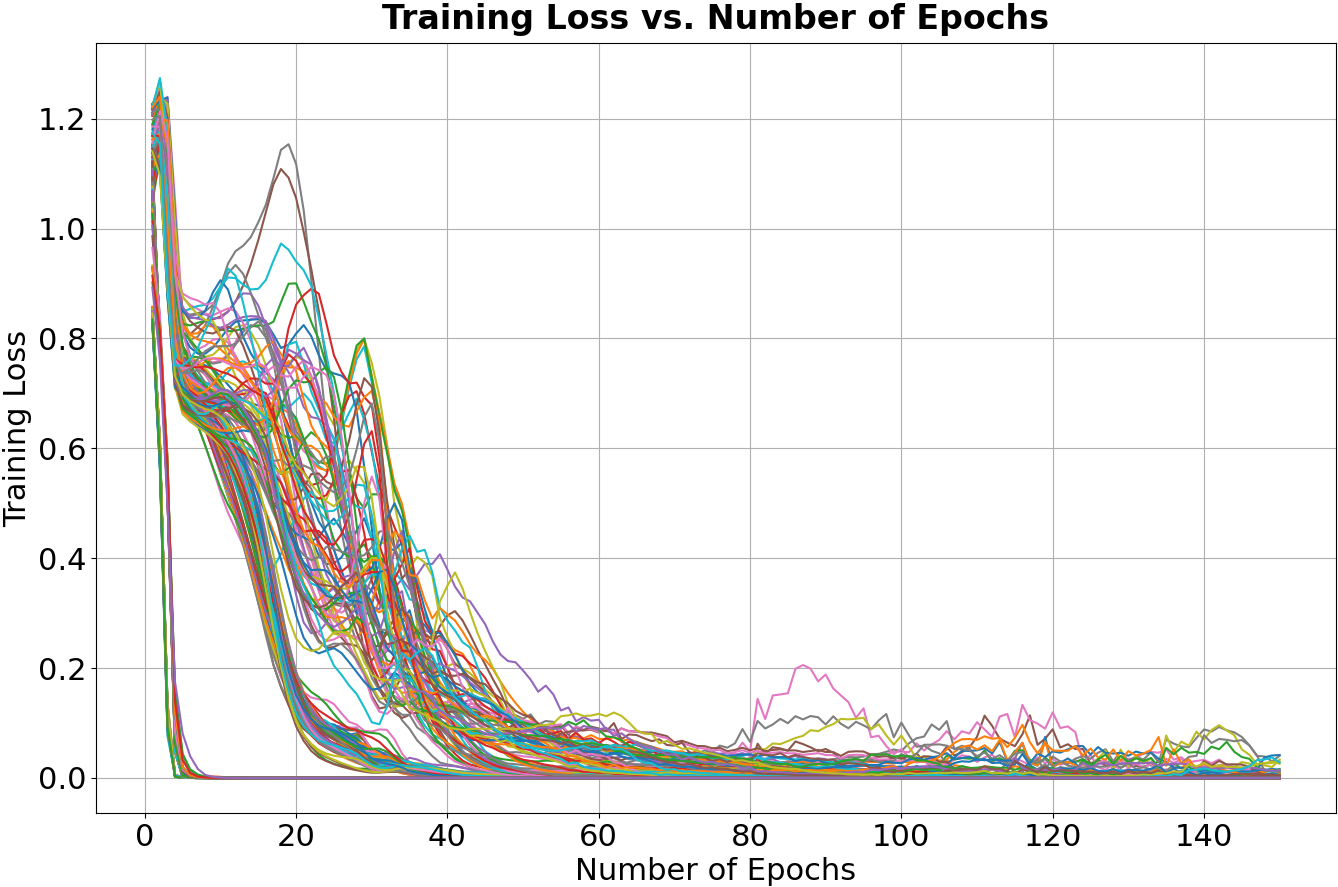}
    \caption{The training loss of each training data instance in the CovidDeep dataset \cite{covidD} across 150 epochs.}
    \label{fig:loss}
\end{figure}

Inspired by a framework called \emph{clustering training losses for label error detection}
(CTRL) \cite{ctrl}, we look into the average training loss of each data instance to evaluate how informative the
instance is. Fig.~\ref{fig:loss} shows a graph of the training loss of each real training data instance in the
CovidDeep dataset \cite{covidD} across 150 epochs. The training losses of some data instances decrease fast and
stay near zero after around 30 epochs. These data instances have low average training loss values and are easy for
the DNN model to learn. Therefore, they do not contain sufficient information to serve as good exemplars for the
model to recall the mission. On the other hand, the training losses of some data remain high
until 60 epochs. These data instances yield higher average training loss values and are more informative about the
mission. Hence, they are good candidates for preservation as exemplars for the model in future replays. We conduct an ablation study for this design decision and give details in Section \ref{sec:ablation1}.

\SetKwComment{Comment}{\# }{}
\begin{algorithm}[t]
\small
\SetAlgoLined
    \SetKwInOut{Input}{Input}
    \SetKwInOut{Output}{Output}
    \Input{accumulated loss matrix ($\mathcal{L}$), training data ($X_{train}$), training data labels ($Y_{train}$), number of epochs ($e$), percentile ($p$), class labels ($L$)}
    \Output{preserved data ($X_{preserved}$), preserved data labels ($Y_{preserved}$)} 
    \BlankLine
    Initialize threshold ($t$)\;
    $\mathcal{L} = \mathcal{L} / e$ \Comment*{average training loss}
    \ForEach {$l \in L$}{
        $t_l = \text{the } p \text{-th percentile in } \mathcal{L}|l$\;
        $(X_{preserved}, Y_{preserved})|l = \text{preserve data instances in } (X_{train}, Y_{train})|l$
        $\text{ whose average training loss values in } \mathcal{L}|l \geq t_l$\;
        $X_{preserved} = \texttt{concatenate} (X_{preserved}, X_{preserved}|l)$\;
        $Y_{preserved} = \texttt{concatenate} (Y_{preserved}, Y_{preserved}|l)$\;
    }
    \caption{Data Preservation} 
    \label{alg:DP}
\end{algorithm}

Algorithm \ref{alg:DP} provides the pseudocode of the DP method in our replay-style CL algorithm.
During the model training process, DOCTOR accumulates the training loss of each real training data instance in a loss
matrix $\mathcal{L}$ after each epoch for the current mission. After training finishes, the DP
method first calculates the average training loss of each data instance by dividing $\mathcal{L}$ by the number of
epochs $e$. Then, for each class label $l$, the algorithm sets a threshold value $t_l$ as the $p$-th percentile of
the average training loss values in $\mathcal{L}|l$. Next, the algorithm preserves the data instances
$X_{preserved}|l$ and their corresponding labels $Y_{preserved}|l$ whose average training loss values in
$\mathcal{L}|l$ are greater than or equal to their corresponding $t_l$. Finally, the preserved data $X_{preserved}$
and their corresponding labels $Y_{preserved}$ are obtained by concatenating all $X_{preserved}|l$ and
$Y_{preserved}|l$, respectively. Therefore, the DP method preserves the most informative subset of
the real training data for the current mission in a stratified manner without excessive computation. The preserved
data can then be used to replay the mission in future CL scenarios. We provide details of an 
ablation study, on the selection of the threshold value $t_l$ (\emph{i.e.}, the $p$-th percentile hyperparameter) to 
preserve different amounts of real training data, in Section \ref{sec:ablation2}.

\subsubsection{Synthetic Data Generation Module}
\label{sec:SDG}
In real-world disease detection application scenarios, models are usually trained with data collected 
from actual patients. The usage and storage of a patient's personal health information are governed by numerous laws and 
restrictions to ensure patient data privacy. Therefore, preserving actual patient data for future use may be prohibited. 
In such cases, DOCTOR adopts an SDG module to generate synthetic data for future replays while preserving data privacy.
The module models the joint multivariate probability distribution of the real training data of the current mission.
It then stores only the learned probability density estimation model for future replays. During the 
training process for a CL scenario, the SDG module generates synthetic data on the fly by sampling from the learned 
distribution. Then, DOCTOR labels the synthetic data with the current DNN model and replays them to retain the learned 
knowledge of the mission. We drew inspiration from a framework called TUTOR \cite{tutor} to build our SDG module. 

The SDG module uses probability density estimation methods to model the PDF of the joint multivariate probability distribution of the given real training data $X_{train}$. We implement the parametric Gaussian mixture model estimation (GMME) method \texttt{GMME()} and the non-parametric KDE method \texttt{KDE()} in our SDG module. It is worth 
noting that previous works \cite{covidD, mhD, tutor, repairs} show that GMME and KDE are good at recreating the probability 
distribution of real data and generating high-quality synthetic data. The generated synthetic data can then replace real 
data in the training process while achieving similar test accuracy.

\SetKwComment{Comment}{\# }{}
\begin{algorithm}[t]
\small
\SetAlgoLined
    \SetKwInOut{Input}{Input}
    \SetKwInOut{Output}{Output}
    \Input{training data ($X_{train}$), validation data ($X_{validation}$), max number of components ($C_{max}$), maximum bandwidth ($h_{max}$), number of synthetic data instances ($count$), trained DNN model ($M$)}
    \Output{synthetic data ($X_{syn}$), synthetic data labels ($Y_{syn}$)} 
    \BlankLine
    $X_{syn\_gmm} = \texttt{GMME}(X_{train}, X_{validation}, C_{max}, count)$\;
    $X_{syn\_kde} = \texttt{KDE}(X_{train}, X_{validation}, h_{max}, count)$\;
    $statistic_{gmm} = \texttt{kstest} (X_{train}, X_{syn\_gmm})$\;
    $statistic_{kde} = \texttt{kstest} (X_{train}, X_{syn\_kde})$\;
    \eIf{$statistic_{gmm} < statistic_{kde}$}{
        $X_{syn} = X_{syn\_gmm}$\;
    }
    {
        $X_{syn} = X_{syn\_kde}$\;
    }
    $Y_{syn} = M(X_{syn})$\;
    \caption{Synthetic Data Generation Module} 
    \label{alg:SDG}
\end{algorithm}

Algorithm \ref{alg:SDG} depicts the top-level pseudocode of the SDG module. First, the module generates synthetic data $X_{syn\_gmm}$ and $X_{syn\_kde}$ from the \texttt{GMME()} and \texttt{KDE()} functions, respectively. Next, to decide which synthetic data to use for future replays, we apply the two-sample Kolmogorov-Smirnov (KS) test \texttt{kstest()} to $X_{syn\_gmm}$ and $X_{syn\_kde}$. The KS test statistic is used to evaluate the closeness of the underlying distributions of the two given samples. A lower KS statistic number shows a higher closeness of the distributions. Therefore, we choose the synthetic data with a lower KS test statistic. Finally, DOCTOR generates labels $Y_{syn}$ for the chosen $X_{syn}$ with the trained DNN model $M$. The synthetic data $X_{syn}$ and their corresponding labels $Y_{syn}$ are then replayed in future CL scenarios to retain the learned knowledge. Next, we elaborate on the details of \texttt{GMME()} and \texttt{KDE()}.

\paragraph{Gaussian Mixture Model Estimation Method}
The GMME method uses a multi-dimensional GMM to model the probability distribution of the given 
real training data $X_{train}$. The distribution is modeled as a mixture of $C$ Gaussian models, as mentioned in 
Section \ref{sec:para}. The PDF of the learned distribution can be expressed through Eq.~(\ref{eq:1}). Although GMME can 
generate high-quality synthetic data to replace real training data, one drawback of this method lies in its learning 
complexity. To address this issue, we use the expectation-maximization (EM) algorithm to determine the GMM 
parameters \cite{tutor}. The EM algorithm iteratively solves this optimization problem in the following two steps 
until convergence:

\begin{itemize}
    \item Expectation-step: Given the current parameters $\theta_i$ of the GMM and observations $X$ from the real 
training data, estimate the probability that the observations belong to each Gaussian model component $c$.
    \item Maximization-step: Find the new parameters $\theta_{i+1}$ that maximizes the expectation derived from 
the expectation-step.
\end{itemize}

\SetKwComment{Comment}{\# }{}
\begin{algorithm}[t]
\small
\SetAlgoLined
    \SetKwInOut{Input}{Input}
    \SetKwInOut{Output}{Output}
    \Input{training data ($X_{train}$), validation data ($X_{validation}$), max number of components ($C_{max}$), number of synthetic data instances ($count$)}
    \Output{GMM synthetic data ($X_{syn\_gmm}$)} 
    \BlankLine
    Initialize the set of numbers of components $\mathcal{C} = \texttt{range}(1, C_{max}, \texttt{step}=1)$\;
    \ForEach {$C \in \mathcal{C}$}{
        $gmm_C = \texttt{GMM.fit}(C, X_{train})$ \Comment*{using EM algorithm}
    }
    $C^* = \argmax\limits_{\mathcal{C}}(gmm_C \texttt{.score}(X_{validation}))$\;
    $gmm^* = \texttt{GMM.fit}(C^*, X_{train})$\;
    $X_{syn\_gmm} = gmm^*\texttt{.sample}(count)$\;
    \caption{Gaussian Mixture Model Estimation} 
    \label{alg:GMM}
\end{algorithm}

Algorithm \ref{alg:GMM} shows the pseudocode of the GMME method \texttt{GMME()}. The algorithm starts by initializing a set $\mathcal{C}$ of candidate numbers for the total number of Gaussian model components ranging from 1 to a user-specified maximum number $C_{max}$. Then, for each candidate number $C$ in $\mathcal{C}$, the algorithm fits a GMM $gmm_C$ to the given real 
training data $X_{train}$ with $C$ Gaussian model components. To prevent the GMM from overfitting on $X_{train}$, the total number of Gaussian models $C$ needs to be chosen appropriately. Hence, the algorithm computes the per-sample average log-likelihood $gmm_C\texttt{.score()}$ of the given real validation data $X_{validation}$ under each $gmm_C$ to evaluate the quality of the model. The number $C$ that maximizes this criterion is chosen as the optimal total number $C^*$. It means that $C^*$ Gaussian models are required to model the probability distribution of $X_{train}$ without overfitting. Finally, the algorithm finalizes the GMM $gmm^*$ with parameter $C^*$ and $X_{train}$. Then, it samples a user-defined number $count$ of synthetic data $X_{syn\_gmm}$ from $gmm^*$.

\paragraph{Kernel Density Estimation Method}
The KDE method approximates the probability distribution of the given real training data as a 
sum of several designated kernel functions. We choose the PDF of the normal distribution as the kernel function. 
Therefore, the approximated PDF $\hat{f}$ of the given real training data at any given point $x$ can be formulated as follows:

\begin{equation*}
    \hat{f}(x) = \frac{1}{Nh}\sum_{i=1}^N\mathcal{N}(\frac{x-x_i}{h}),
\end{equation*}

\noindent where $N$ is the total number of the given real training data, $x_i$ is the i-th sample of the data, $\mathcal{N}$ denotes the normal distribution, and $h$ represents the kernel bandwidth. As mentioned in Section \ref{sec:non-para}, the kernel bandwidth $h$ is an important hyperparameter that affects the bias-variance trade-off of the estimator. Therefore, we use a validation set held out from the real data to find the optimal bandwidth value.

\SetKwComment{Comment}{\# }{}
\begin{algorithm}[t]
\small
\SetAlgoLined
    \SetKwInOut{Input}{Input}
    \SetKwInOut{Output}{Output}
    \Input{training data ($X_{train}$), validation data ($X_{validation}$), maximum bandwidth ($h_{max}$), number of synthetic data instances ($count$)}
    \Output{KDE synthetic data ($X_{syn\_kde}$)} 
    \BlankLine
    Initialize the set of bandwidths $\mathcal{H} = \texttt{range}(0.05, h_{max}, \texttt{step}=0.05)$\;
    \ForEach {$h \in \mathcal{H}$}{
        $kde_h = \texttt{KDE.fit}(h, X_{train})$\;
    }
    $h^* = \argmax\limits_{\mathcal{H}}(kde_h \texttt{.score}(X_{validation}))$\;
    $kde^* = \texttt{KDE.fit}(h^*, X_{train})$\;
    $X_{syn\_kde} = kde^*\texttt{.sample}(count)$\;
    \caption{Kernel Density Estimation} 
    \label{alg:KDE}
\end{algorithm}

Algorithm \ref{alg:KDE} shows the pseudocode of the KDE method \texttt{KDE()}. First, the algorithm initializes a set
$\mathcal{H}$ of candidate values for the kernel bandwidth ranging from 0.05 to a user-specified maximum bandwidth
$h_{max}$. Then, for each candidate bandwidth $h$ in $\mathcal{H}$, the algorithm fits a KDE model $kde_h$ on the given
real training data $X_{train}$ with bandwidth $h$. Next, as discussed earlier, the algorithm finds the appropriate
kernel bandwidth value for the final KDE model to prevent overfitting. The algorithm computes the total log-likelihood
$kde_h.\texttt{score()}$ of the given real validation data $X_{validation}$ under each $kde_h$ to evaluate the model's
quality. The value $h$ that maximizes this criterion is designated as the final kernel bandwidth $h^*$. It represents
the bandwidth required to model the probability distribution of $X_{train}$ without overfitting. Finally, the algorithm
fits the final KDE model $kde^*$ with parameter $h^*$ and $X_{train}$, and samples a user-defined number $count$ of synthetic data $X_{syn\_kde}$ from $kde^*$.

\section{Experimental Setup}
\label{sec:setup}
In this section, we provide details of the experimental setup. First, we introduce the datasets that are used in our
experiments. Next, we describe the preprocessing processes that are applied to these datasets. Finally, we provide implementation details of our DOCTOR framework.

\subsection{Datasets}
\label{sec:datasets}
To evaluate the efficacy of DOCTOR in continually learning various disease detection missions, 
we conduct experiments with three different disease datasets: CovidDeep \cite{covidD}, DiabDeep \cite{diabD}, and 
MHDeep \cite{mhD}. The data collection of these datasets and the experimental procedures were approved by the 
Institutional Review Board of Princeton University. The efficacy of these datasets has been demonstrated in the 
literature \cite{covidD, diabD, mhD}.

The CovidDeep dataset includes physiological signals and responses to a simple questionnaire 
acquired from 38 healthy individuals, 30 asymptomatic patients, and 32 symptomatic patients at San Matteo Hospital 
in Pavia, Italy. While the physiological signals were recorded as continuous values, the responses to the 
questionnaire had Boolean values. The data were collected with commercially available WMSs and devices, including 
an Empatica E4 smartwatch, a pulse oximeter, and a blood pressure monitor. A series of experiments was described in 
\cite{covidD} to identify which data features are relevant to detecting the COVID-19 virus in patients. 
Table~\ref{tbl:CDFeatures} shows the data features that were found to be beneficial in obtaining a high 
disease detection accuracy.

\begin{table}[t]
    \caption{Data Features Collected in the CovidDeep Dataset}
    \centering
    \resizebox{0.65\linewidth}{!}{
    \begin{tabular}{lcc}
    \toprule
    Data Features & Data Source & Data Type \\
    \midrule
    Galvanic Skin Response ($\mu S$) & \\
    Skin Temperature ($^\circ C$) & Smartwatch & Continuous \\
    Inter-beat Interval ($ms$) & \\
    \midrule
    Oxygen Saturation ($\%$) & Pulse Oximeter & Continuous \\
    Systolic Blood Pressure (mmHg) & Blood Pressure Monitor & Continuous \\
    Diastolic Blood Pressure (mmHg) & Blood Pressure Monitor & Continuous \\
    \midrule
    Immune-compromised & \\
    Chronic Lung Disease & \\
    Shortness of Breath & \\
    Cough & \\
    Fever & \\
    Muscle Pain & Questionnaire & Boolean \\
    Chills & \\
    Headache & \\
    Sore Throat & \\
    Smell/Taste Loss & \\
    Diarrhea & \\
    \bottomrule
    \end{tabular}
    }
    \label{tbl:CDFeatures}
\end{table}

\begin{table}[t]
    \caption{Data Features Collected in the DiabDeep Dataset and the MHDeep Dataset}
    \centering
    \resizebox{0.55\linewidth}{!}{
    \begin{tabular}{lcc}
    \toprule
    Data Features & Data Source & Data Type \\
    \midrule
    Galvanic Skin Response ($\mu S$) & \\
    Skin Temperature ($^\circ C$) & \\
    Acceleration ($x, y, z$) & Smart Watch & Continuous \\
    Inter-beat Interval ($ms$) & \\
    Blood Volume Pulse & \\
    \midrule
    Humidity & \\
    Ambient Illuminance & \\
    Ambient Light Color Spectrum & \\
    Ambient Temperature & \\
    Gravity ($x, y, z$) & \\
    Angular Velocity ($x, y, z$) & \\
    Orientation ($x, y, z$) & Smart Phone & Continuous \\
    Acceleration ($x, y, z$) & \\
    Linear Acceleration ($x, y, z$) & \\
    Air Pressure & \\
    Proximity & \\
    Wi-Fi Radiation Strength & \\
    Magnetic Field Strength & \\
    \bottomrule
    \end{tabular}
    }
    \label{tbl:DDFeatures}
\end{table}

The DiabDeep dataset contains various data features obtained from 25 non-diabetic individuals, 
14 Type-I diabetic patients, and 13 Type-II diabetic patients. All the data features were recorded as continuous 
values. The physiological signals of the participants were collected with an Empatica E4 smartwatch. Additional 
electromechanical and ambient environmental data were recorded with a Samsung Galaxy S4 smartphone. These 
additional data were proven to provide information that offers diagnostic insights through user habit tracking 
\cite{additional_data}. For example, they can assist in sensing body movement and calibrating physiological signal 
collection. Table~\ref{tbl:DDFeatures} lists the data features that were collected in this dataset.

The MHDeep dataset contains the same physiological signals and additional continuous data 
features as those collected in the DiabDeep dataset. The data were collected from 23 healthy participants, 23 
participants with bipolar disorder, 10 participants with major depressive disorder, and 16 participants with 
schizoaffective disorder at the Hackensack Meridian Health Carrier Clinic, Belle Mead, New Jersey. As before, 
the data were collected with an Empatica E4 smartwatch and a Samsung Galaxy S4 smartphone. The data features 
included in this dataset are shown in Table~\ref{tbl:DDFeatures} as well.

\subsection{Dataset Preprocessing}
\label{sec:preprocessing}
We preprocess all three datasets before using them in our experiments. To avoid time correlation between
adjacent data windows, We first synchronize and window the data streams by dividing data into 15-second
windows with 15-second shifts in between. Each 15-second window of data constitutes one data instance. Next,
we flatten and concatenate the data within the same time window from the WMSs and smartphones. Then, we
concatenate the sequential time series data with the responses to the questionnaire for the CovidDeep data.
This results in a total of 14047 data instances with 155 features each for the CovidDeep dataset, a total of
20957 data instances with 4485 features each for the DiabDeep dataset, and a total of 27082 data instances
with 4485 features each for the MHDeep dataset. Following this, we perform min-max
normalization on the feature data in all datasets to scale them into the range between 0 and 1. This prevents 
features with a wider range of values from overshadowing those with a narrower range.

To align the input feature dimensions of all three datasets for CL experiments, we apply principal component analysis to the DiabDeep and MHDeep datasets to reduce their dimensionality from 4485 to 155. First, we standardize the feature data in the datasets and compute the covariance matrix of the features. Then, we perform eigendecomposition to find the eigenvectors and eigenvalues of the covariance matrix to identify the principal components. Next, we order the eigenvectors in descending order based on the magnitude of their corresponding eigenvalues. Finally, we construct a projection matrix to select the top 155 principal components. Then, we recast the data along the principal component axes to retrieve the 155-dimensional feature data for these datasets. Subsequently, we partition the datasets into training, validation, and test sets based on different CL scenarios with no time overlap (see details in Section \ref{sec:results}).

Last but not least, we apply the Synthetic Minority Oversampling Technique (SMOTE) \cite{SMOTE} to the 
partitioned training datasets to counteract the data imbalance issue within each dataset. We start by selecting a
random data instance $A$ in a minority class and finding its five nearest neighbors in that class. Then, we randomly
select one nearest neighbor $B$ from the five and draw a line segment between $A$ and $B$ in the feature space. Finally, a synthetic data instance is generated at a randomly selected point on the line segment between $A$ and $B$. We then repeat this process until we obtain a balanced number of data instances in all classes in each partitioned training set.

\subsection{Implementation Details}
\label{sec:implement}
DOCTOR exploits a single DNN model and the proposed replay-style CL algorithm to learn various disease detection tasks incrementally in the tabular data domain. Other generic DNN models that perform classification on sequential time series data can be employed in our framework, such as RNNs and LSTM networks (see an ablation study in Section \ref{sec:ablation4}). However, these networks generally require more training data, computational resources, and time for training. Thus, we choose to implement a multilayer perceptron (MLP) model in our framework. 

\begin{figure}[t]
    \centering
    \includegraphics[width=0.6\linewidth]{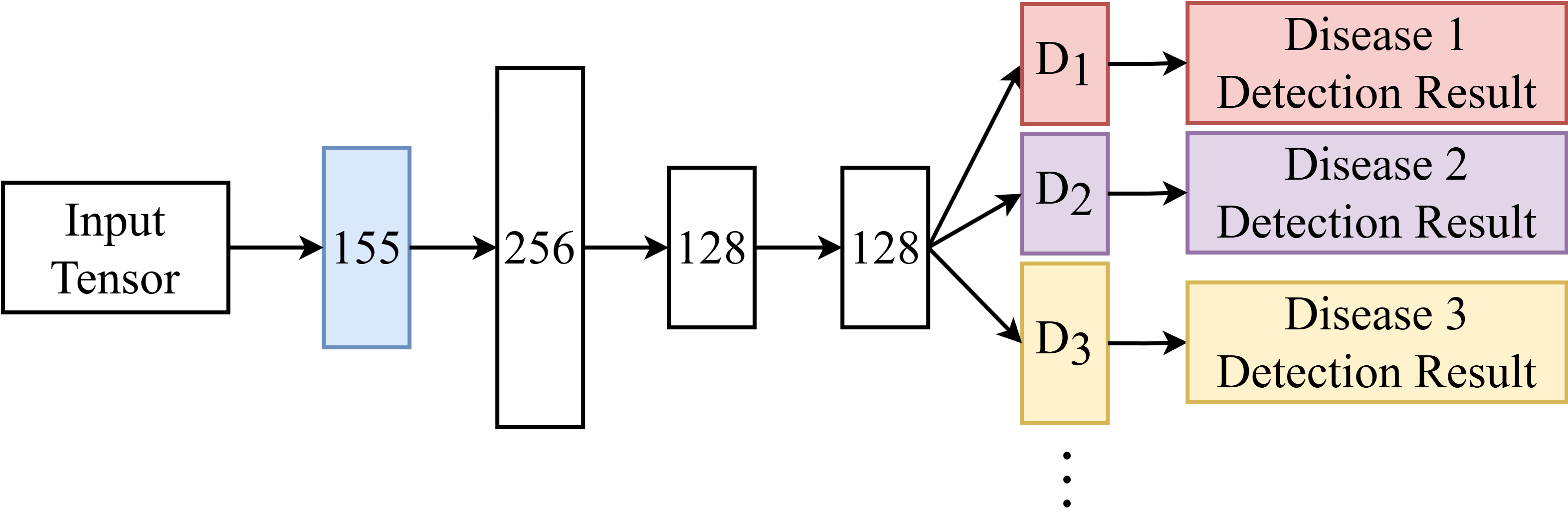}
    \caption{The DNN architecture in DOCTOR. D$_i$ represents the output head of the $i$-th disease detection task.}
    \label{fig:dnn}
\end{figure}

We train and evaluate various MLP models with different numbers of layers and numbers of neurons
per layer on the validation sets of the three datasets. We find that a four-layer architecture performs the best in general. Fig.~\ref{fig:dnn} shows the architecture of the final MLP model. It has an input layer with 155 neurons to align with the dimensionality of input features. Subsequently, it has three hidden layers with 256, 128, and 128 neurons in each layer, respectively. Finally, it incorporates a multi-headed architecture in the output layer to accommodate each learned disease detection task. Each output head has a varying number of neurons corresponding to the number of classification classes in that task. We adopt the rectified linear unit (ReLU) as the nonlinear activation function in the hidden layers and the softmax function for each head in the output layer to generate detection results.

We use the stochastic gradient descent (SGD) optimizer with a momentum of 0.9 in our experiments and initialize the
learning rate to 0.005. We set the batch size to 128 for training, where data are drawn evenly from the current mission
and each previous mission. We train the MLP model for 300 epochs. For the DP method, we set the threshold value as the
70th percentile (see details in Section \ref{sec:ablation}) of the average training loss values of each classification class. 
We set the maximum number of components in the SDG module to 50 for the GMME method and the maximum bandwidth to 0.5 for the KDE method. Finally, we sample the same number of synthetic data instances from the module as the number of real training data instances. 

We implement DOCTOR with PyTorch and perform the experiments on an NVIDIA A100 GPU. We employ CUDA and cuDNN libraries to accelerate the experiments.

\section{Experimental Results}
\label{sec:results}
In this section, we present the experimental results of our DOCTOR framework. We compare
DOCTOR with a baseline framework and two other common approaches in the experiments, including a naive
fine-tuning and an ideal joint-training framework. The baseline framework trains the MLP model with the first mission
only and uses the performance of the model as the baseline for CL. The naive fine-tuning framework just naively fine-tunes the MLP model with data from the new mission without adopting any CL algorithm. The ideal joint-training framework represents the ideal scenario where the MLP model has access to and is jointly trained with all data from the new and previous missions. On the other hand, to compare DOCTOR with existing works, we surveyed various CL frameworks. However, existing architecture-based and replay-based CL frameworks target the image classification domain \cite{review, overview, survey_a, survey_b, survey_c, survey_d, l2p, CLwVis} and their methods are not directly 
applicable to tabular data, where the data modality is significantly different. Therefore, we implement the regularization-based CL framework, LwF \cite{lwf}, whose method is transferable to the tabular data domain and extensible to all three CL scenarios with a similar multi-headed architecture as DOCTOR.

We evaluate DOCTOR's efficacy under three different CL settings: domain-incremental, class-incremental, and 
task-incremental. In addition, we examine the performance of all three replay-style CL methods 
in DOCTOR: DP, SDG with GMME, and SDG with KDE. We conduct experiments with the CovidDeep, 
DiabDeep, and MHDeep datasets introduced in Section \ref{sec:datasets}. We report the 
performance of all frameworks with the evaluation metrics mentioned in Section \ref{sec:metrics}, including 
average accuracy ($Acc_{avg}$), average F1-score (F1-score$_{avg}$), and BWT. In addition, we report the test 
accuracy ($Acc$) for each learned mission and the required buffer size in megabytes (MB) for all frameworks. 
Moreover, we record the KS test statistic (described in Section \ref{sec:SDG}) for the two SDG methods in DOCTOR. 
Finally, we demonstrate the potential of the multi-disease detection application of DOCTOR and conduct an ablation study.

In all frameworks, the MLP models inherit the weights from the baseline framework and then learn subsequent missions with their associated CL methods. We repeat each CL experiment three times for all frameworks and report the average values for all evaluation metrics. All experimental results are reported on the \emph{real test data} held out from their corresponding datasets.

\begin{table*}[t]
    \caption{Domain-incremental CL Experimental Results}
    \centering
    \resizebox{1\linewidth}{!}{
    \begin{tabular}{crlccccccc}
    \toprule
    Datasets & \multicolumn{2}{c}{Frameworks} & $M_1~Acc$ & $M_2~Acc$ & $Acc_{avg}$ & F1-score$_{avg}$ & BWT & KS Test Statistic & Buffer Size (MB) \\
    \midrule
    \multirow{7}{*}{CovidDeep} & \multicolumn{2}{c}{Ideal Joint-training} & 0.992 & 0.998 & 0.995 & 0.998 & 0.008 & - & 5.8 \\
     & \multirow{3}{*}{DOCTOR} & - DP & 0.979 & 1.000 & \textbf{0.990} & 0.994 & \textbf{-0.007} & - & 1.7 \\
     & & - SDG with GMME & 0.978 & 0.993 & 0.986 & \textbf{0.995} & \textbf{-0.007} & 0.083 & 0 \\
     & & - SDG with KDE & 0.978 & 1.000 & 0.989 & \textbf{0.995} & \textbf{-0.007} & 0.079 & 0 \\
     & \multicolumn{2}{c}{LwF} & 0.883 & 1.000 & 0.941 & 0.968 & -0.097 & - & 0 \\
     & \multicolumn{2}{c}{Naive Fine-tuning} & 0.767 & 1.000 & 0.884 & 0.929 & -0.217 & - & 0 \\
     \cmidrule{2-10}
     & \multicolumn{2}{c}{Baseline} & 0.984 & 0.755 & 0.870 & 0.906 & - & - & 0 \\
    \midrule
    \multirow{7}{*}{DiabDeep} & \multicolumn{2}{c}{Ideal Joint-training} & 0.920 & 0.991 & 0.955 & 0.960 & -0.002 & - & 11.5 \\
     & \multirow{3}{*}{DOCTOR} & - DP & 0.911 & 0.995 & 0.954 & 0.957 & -0.011 & - & 3.4 \\
     & & - SDG with GMME & 0.907 & 0.963 & 0.935 & 0.944 & -0.014 & 0.215 & 0 \\
     & & - SDG with KDE & 0.918 & 0.995 & \textbf{0.957} & \textbf{0.961} & \textbf{-0.004} & 0.050 & 0 \\
     & \multicolumn{2}{c}{LwF} & 0.751 & 0.980 & 0.866 & 0.876 & -0.185 & - & 0 \\
     & \multicolumn{2}{c}{Naive Fine-tuning} & 0.445 & 1.000 & 0.723 & 0.796 & -0.477 & - & 0 \\
     \cmidrule{2-10}
     & \multicolumn{2}{c}{Baseline} & 0.923 & 0.364 & 0.643 & 0.664 & - & - & 0 \\
    \midrule
    \multirow{7}{*}{MHDeep} & \multicolumn{2}{c}{Ideal Joint-training} & 0.834 & 0.942 & 0.888 & 0.979 & -0.006 & - & 13.6 \\
     & \multirow{3}{*}{DOCTOR} & - DP & 0.819 & 0.954 & \textbf{0.887} & \textbf{0.977} & -0.013 & - & 4.1 \\
     & & - SDG with GMME & 0.827 & 0.894 & 0.860 & 0.958 & -0.006 & 0.041 & 0 \\
     & & - SDG with KDE & 0.828 & 0.942 & 0.885 & 0.976 & \textbf{-0.005} & 0.036 & 0 \\
     & \multicolumn{2}{c}{LwF} & 0.577 & 0.936 & 0.757 & 0.883 & -0.262 & - & 0 \\
     & \multicolumn{2}{c}{Naive Fine-tuning} & 0.280 & 0.986 & 0.633 & 0.794 & -0.554 & - & 0 \\
     \cmidrule{2-10}
     & \multicolumn{2}{c}{Baseline} & 0.833 & 0.131 & 0.482 & 0.757 & - & - & 0 \\
    \bottomrule
    \end{tabular}
    }
    \label{tbl:DIL}
\end{table*}

\subsection{Domain-incremental Continual Learning Scenario}
\label{sec:DIL}
In the domain-incremental CL experiments, we assume two missions where data from two different distributions for the
same disease detection task become available sequentially. For all three datasets, we first randomly split the patients
into two missions (groups) in a stratified fashion, where Mission 1 ($M_1$) has 80\% of the patients from each class
and Mission 2 ($M_2$) has the remaining 20\%. Then, within each mission, we take the first 70\%, the next 10\%, and the last 20\% of each patient's sequential time series data to construct the training, validation, and test sets with no time overlap. Subsequently, we apply SMOTE to the training sets in both missions to counteract the data imbalance issue.

In the domain-incremental scenario, new missions just contain data from different probability distributions but with the 
same classification classes as the given task. Hence, the MLP models do not need to expand their output layers to
accommodate new missions. Table~\ref{tbl:DIL} shows the experimental results for all frameworks in the
domain-incremental CL scenario. As we can see from the table, for all three datasets, the MLP model suffers from
significant performance deterioration for $M_1$ when the naive fine-tuning framework is used, due to catastrophic forgetting. This results in poor $Acc_{avg}$, F1-score$_{avg}$, and BWT compared to the other frameworks. DOCTOR far outperforms the 
naive fine-tuning and LwF frameworks, and it achieves very competitive $Acc_{avg}$, F1-score$_{avg}$, and BWT relative to the ideal joint-training framework for all three datasets. Moreover, DOCTOR even outperforms the ideal scenario in some cases due to the generalization gained through learning from synthetic data. Next, we 
examine the KS test statistics when DOCTOR employs generative replay with the SDG module. When GMME and KDE both 
yield similar KS test statistics for the CovidDeep and MHDeep datasets, DOCTOR achieves a very similar test 
accuracy for $M_1$ with the synthetic data generated using each method. However, for the DiabDeep dataset, the 
KDE method results in a lower KS test statistic and hence a higher test accuracy for $M_1$. This validates our 
design decision described in Section \ref{sec:SDG} in choosing the estimation method that yields a lower KS test 
statistic for generative replay.

In summary, DOCTOR achieves a 0.990 $Acc_{avg}$, a 0.995 F1-score$_{avg}$, and a -0.007 BWT on the CovidDeep dataset, a 0.957 $Acc_{avg}$, a 0.961 F1-score$_{avg}$, and a -0.004 BWT on the DiabDeep dataset, and a 0.887 $Acc_{avg}$, a 0.977 F1-score$_{avg}$, and a -0.005 BWT on the MHDeep dataset. This demonstrates that DOCTOR can incrementally adapt to data from new distributions while maintaining the knowledge learned from previous ones. The best DOCTOR results are shown in bold.

\subsection{Class-incremental Continual Learning Scenario}
\label{sec:CIL}
In the class-incremental CL experiments, we assume two missions where $M_2$ contains new classification
classes not seen in $M_1$ for the same disease detection task. For CovidDeep, we include only the healthy
individuals and symptomatic patients in $M_1$, whereas $M_2$ only contains the asymptomatic patients. For
DiabDeep, $M_1$ includes the healthy individuals and Type-I diabetic patients, whereas $M_2$ comprises solely the Type-II diabetic patients. For MHDeep, we include the healthy participants and participants with major depressive disorder in $M_1$, whereas $M_2$ consists of participants with bipolar depressive disorder and schizoaffective disorder. As before, within each mission, we prepare the training, validation, and test sets by taking the first 70\%, the next 10\%, and the last 20\% of the sequential time series data instances from each patient with no time overlap. Similarly, we apply SMOTE to the training sets in both missions to address the data imbalance issue.

In the class-incremental CL scenario, the MLP models in all frameworks except for the baseline framework need to add more output neurons to accommodate the unseen classification classes. Therefore, in those frameworks, they first inherit the weights 
of their MLP models from the baseline framework and then add the same number of new output neurons as the number of unseen classes in $M_2$. Then, the MLP models are trained with various CL methods corresponding to their frameworks.

Table~\ref{tbl:CIL} presents the experimental results for all frameworks in the class-incremental CL
scenario. As can again be seen from the table, due to catastrophic forgetting, the MLP model suffers from significant performance degradation on $M_1$ when the naive fine-tuning framework is used in all three datasets. However, DOCTOR significantly outperforms the naive fine-tuning and LwF frameworks. Moreover, due to fine-tuning with the most informative preserved data and the generalization gained through learning from synthetic data, it even achieves higher $Acc_{avg}$ and BWT and a competitive F1-score$_{avg}$ relative to the ideal joint-training framework for all three datasets. On the other hand, as shown in the table, the estimation method that yields a lower KS 
test statistic results in a better test accuracy for $M_1$ for all datasets. This again validates the design 
decision we made regarding the KS test statistic in Section \ref{sec:SDG}.

\begin{table*}[t]
    \caption{Class-incremental CL Experimental Results}
    \centering
    \resizebox{1\linewidth}{!}{
    \begin{tabular}{crlccccccc}
    \toprule
    Datasets & \multicolumn{2}{c}{Frameworks} & $M_1~Acc$ & $M_2~Acc$ & $Acc_{avg}$ & F1-score$_{avg}$ & BWT & KS Test Statistic & Buffer Size (MB) \\
    \midrule
    \multirow{7}{*}{CovidDeep} & \multicolumn{2}{c}{Ideal Joint-training} & 0.989 & 0.982 & 0.985 & 0.993 & -0.009 & - & 5.0 \\
     & \multirow{3}{*}{DOCTOR} & - DP & 0.985 & 0.987 & \textbf{0.986} & \textbf{0.991} & \textbf{-0.008} & - & 1.5 \\
     & & - SDG with GMME & 0.937 & 0.980 & 0.958 & 0.966 & -0.056 & 0.166 & 0 \\
     & & - SDG with KDE & 0.977 & 0.977 & 0.977 & 0.984 & -0.016 & 0.081 & 0 \\
     & \multicolumn{2}{c}{LwF} & 0 & 1.000 & 0.500 & 0.837 & -0.998 & - & 0 \\
     & \multicolumn{2}{c}{Naive Fine-tuning} & 0 & 1.000 & 0.500 & 0.837 & -0.992 & - & 0 \\
     \cmidrule{2-10}
     & \multicolumn{2}{c}{Baseline} & 0.993 & 0 & 0.496 & 0.751 & - & - & 0 \\
    \midrule
    \multirow{7}{*}{DiabDeep} & \multicolumn{2}{c}{Ideal Joint-training} & 0.881 & 0.926 & 0.903 & 0.932 & -0.048 & - & 9.2 \\
     & \multirow{3}{*}{DOCTOR} & - DP & 0.847 & 0.936 & 0.891 & 0.914 & -0.081 & - & 2.8 \\
     & & - SDG with GMME & 0.903 & 0.938 & \textbf{0.921} & 0.925 & \textbf{-0.025} & 0.063 & 0 \\
     & & - SDG with KDE & 0.879 & 0.937 & 0.908 & \textbf{0.929} & -0.050 & 0.211 & 0 \\
     & \multicolumn{2}{c}{LwF} & 0 & 1.000 & 0.500 & 0.647 & -0.934 & - & 0 \\
     & \multicolumn{2}{c}{Naive Fine-tuning} & 0 & 1.000 & 0.500 & 0.647 & -0.928 & - & 0 \\
     \cmidrule{2-10}
     & \multicolumn{2}{c}{Baseline} & 0.928 & 0 & 0.464 & 0.702 & - & - & 0 \\
    \midrule
    \multirow{7}{*}{MHDeep} & \multicolumn{2}{c}{Ideal Joint-training} & 0.891 & 0.897 & 0.893 & 0.985 & -0.055 & - & 8.3 \\
     & \multirow{3}{*}{DOCTOR} & - DP & 0.886 & 0.888 & 0.887 & \textbf{0.983} & -0.059 & - & 2.5 \\
     & & - SDG with GMME & 0.944 & 0.846 & \textbf{0.895} & 0.973 & \textbf{-0.001} & 0.042 & 0 \\
     & & - SDG with KDE & 0.897 & 0.887 & 0.892 & 0.981 & -0.048 & 0.209 & 0 \\
     & \multicolumn{2}{c}{LwF} & 0.012 & 0.894 & 0.453 & 0.775 & -0.931 & - & 0 \\
     & \multicolumn{2}{c}{Naive Fine-tuning} & 0 & 0.895 & 0.448 & 0.775 & -0.945 & - & 0 \\
     \cmidrule{2-10}
     & \multicolumn{2}{c}{Baseline} & 0.945 & 0 & 0.472 & 0.494 & - & - & 0 \\
    \bottomrule
    \end{tabular}
    }
    \label{tbl:CIL}
\end{table*}

To summarize, DOCTOR achieves a 0.986 $Acc_{avg}$, a 0.991 F1-score$_{avg}$, and a -0.008 BWT on the CovidDeep dataset, a 0.921 $Acc_{avg}$, a 0.929 F1-score$_{avg}$, and a -0.025 BWT on the DiabDeep dataset, and a 0.895 $Acc_{avg}$, a 0.983 F1-score$_{avg}$, and a -0.001 BWT on the MHDeep dataset. This demonstrates the efficacy of DOCTOR in learning new classification classes while preserving prior knowledge. Again, the best DOCTOR results are shown in bold.

\subsection{Task-incremental Continual Learning}
\label{sec:TIL}
In the task-incremental CL experiments, we assign three missions where DOCTOR learns three different disease
detection tasks incrementally with the three disease datasets. First, we let DOCTOR learn to detect the
COVID-19 virus among healthy, symptomatic, and asymptomatic patients using the CovidDeep dataset ($M_1$).
Next, we have DOCTOR learn to differentiate between healthy, Type-I diabetic, and Type-II diabetic patients using the
DiabDeep dataset ($M_2$). Finally, we have DOCTOR learn to recognize participants that are healthy or suffer from bipolar,
major depressive, or schizoaffective disorder using the MHDeep dataset ($M_3$). For each mission, we construct the
training, validation, and test sets by taking the first 70\%, the next 10\%, and the last 20\% of the sequential time
series data instances from each patient in the disease dataset with no time overlap. Finally, we apply SMOTE to the training sets in all three missions to handle the data imbalance issue in the datasets.

In the baseline framework, the MLP model only contains a single detection head in the output layer and
is only trained with CovidDeep data. The CovidDeep detection head has the same number of output neurons as the number of classes in the CovidDeep detection task. When $M_2$ arrives, all the other frameworks first inherit the weights of their MLP models from the baseline framework. Then, they generate a new detection head parallel to the CovidDeep detection head in the output layer. The new head has the same number of output neurons as the number of classes in the DiabDeep detection task. Then, the MLP models are trained accordingly with respect to their CL frameworks.

When $M_3$ becomes available, all frameworks except the baseline inherit the weights of their MLP models from their previous states, respectively. Then, they generate a new detection head in parallel with the CovidDeep and DiabDeep detection heads in the output layer. The newly generated head contains the same number of neurons as the number of classes in the MHDeep detection task. Then, the MLP models learn the new detection task according to their associated CL frameworks.

Table~\ref{tbl:TIL} shows the experimental results for all frameworks in the task-incremental CL
scenario. As shown in the table, due to catastrophic forgetting, the naive fine-tuning 
framework consistently suffers from performance deterioration in the previously learned detection tasks after 
learning a new task. Our DOCTOR framework consistently outperforms both the naive fine-tuning and LwF frameworks. 
It achieves very competitive performance relative to the ideal joint-training framework even after incrementally 
learning multiple disease detection tasks. Moreover, DOCTOR is able to maintain high test accuracy for all the 
learned detection tasks in the task-incremental CL scenario. Both GMME and KDE methods yield 
similar KS test statistics in this scenario. Thus, DOCTOR achieves very similar test accuracy for 
$M_1$ and $M_2$ with the synthetic data generated from both estimation methods.

\begin{table*}[t]
    \caption{Task-incremental CL Experimental Results}
    \centering
    \resizebox{1\linewidth}{!}{
    \begin{tabular}{crlcccccccc}
    \toprule 
    Datasets & \multicolumn{2}{c}{Frameworks} & $M_1~Acc$ & $M_2~Acc$ & $M_3~Acc$ & $Acc_{avg}$ & F1-score$_{avg}$ & BWT & KS Test Statistic & Buffer Size (MB) \\
    \midrule
    MHDeep & \multicolumn{2}{c}{Ideal Joint-training} & 0.991 & 0.929 & 0.869 & 0.930 & 0.970 & 0.001 & - & 21.1 \\
    $\uparrow$ & \multirow{3}{*}{DOCTOR} & - DP & 0.989 & 0.937 & 0.858 & \textbf{0.928} & \textbf{0.969} & \textbf{0.002} & - & 6.3 \\
    \multirow{2}{*}{DiabDeep} & & - SDG with GMME & 0.986 & 0.921 & 0.855 & 0.920 & 0.964 & -0.007 & 0.036 & 0 \\
    \multirow{2}{*}{$\uparrow$} & & - SDG with KDE & 0.984 & 0.931 & 0.854 & 0.923 & 0.967 & -0.002 & 0.041 & 0 \\
     & \multicolumn{2}{c}{LwF} & 0.541 & 0.891 & 0.852 & 0.761 & 0.907 & -0.243 & - & 0 \\
    CovidDeep & \multicolumn{2}{c}{Naive Fine-tuning} & 0.623 & 0.477 & 0.847 & 0.649 & 0.773 & -0.408 & - & 0 \\
    \midrule
    \multirow{2}{*}{DiabDeep} & \multicolumn{2}{c}{Ideal Joint-training} & 0.991 & 0.931 & - & 0.961 & 0.964 & 0.005 & - & 7.3 \\
     & \multirow{3}{*}{DOCTOR} & - DP & 0.993 & 0.932 & - & \textbf{0.962} & \textbf{0.964} & \textbf{0.002} & - & 2.2 \\
    \multirow{2}{*}{$\uparrow$} & & - SDG with GMME & 0.985 & 0.930 & - & 0.958 & 0.962 & -0.006 & 0.081 & 0 \\
     & & - SDG with KDE & 0.984 & 0.929 & - & 0.957 & 0.962 & -0.006 & 0.079 & 0 \\
    \multirow{2}{*}{CovidDeep} & \multicolumn{2}{c}{LwF} & 0.603 & 0.937 & - & 0.770 & 0.901 & -0.378 & - & 0 \\
     & \multicolumn{2}{c}{Naive Fine-tuning} & 0.761 & 0.925 & - & 0.843 & 0.874 & -0.230 & - & 0 \\
    \midrule
     CovidDeep & \multicolumn{2}{c}{Baseline} & 0.991 & - & - & 0.991 & 0.996 & - & - & 0 \\
    \bottomrule
    \end{tabular}
    }
    \label{tbl:TIL}
\end{table*}

DOCTOR achieves a 0.962 $Acc_{avg}$, a 0.964 F1-score$_{avg}$, and a 0.002 BWT after continually learning two disease
detection tasks, and a 0.928 $Acc_{avg}$, a 0.969 F1-score$_{avg}$, and a 0.002 BWT after consecutively learning to
detect three diseases. This demonstrates DOCTOR's efficacy in the task-incremental scenario. The best DOCTOR results
are again shown in bold.

The initial model size of DOCTOR after learning the CovidDeep dataset is 342KB. The size increases to 344KB after learning to detect the DiabDeep dataset due to the addition of three output neurons in the new detection head. Finally, the size increases to 346KB after learning the MHDeep dataset due to the addition of another detection head with four output neurons. This allows DOCTOR to fit in various edge devices and makes it a promising application for efficient and out-of-clinic disease detection. 

\begin{figure}[t]
    \centering    
    \includegraphics[width=0.6\linewidth]{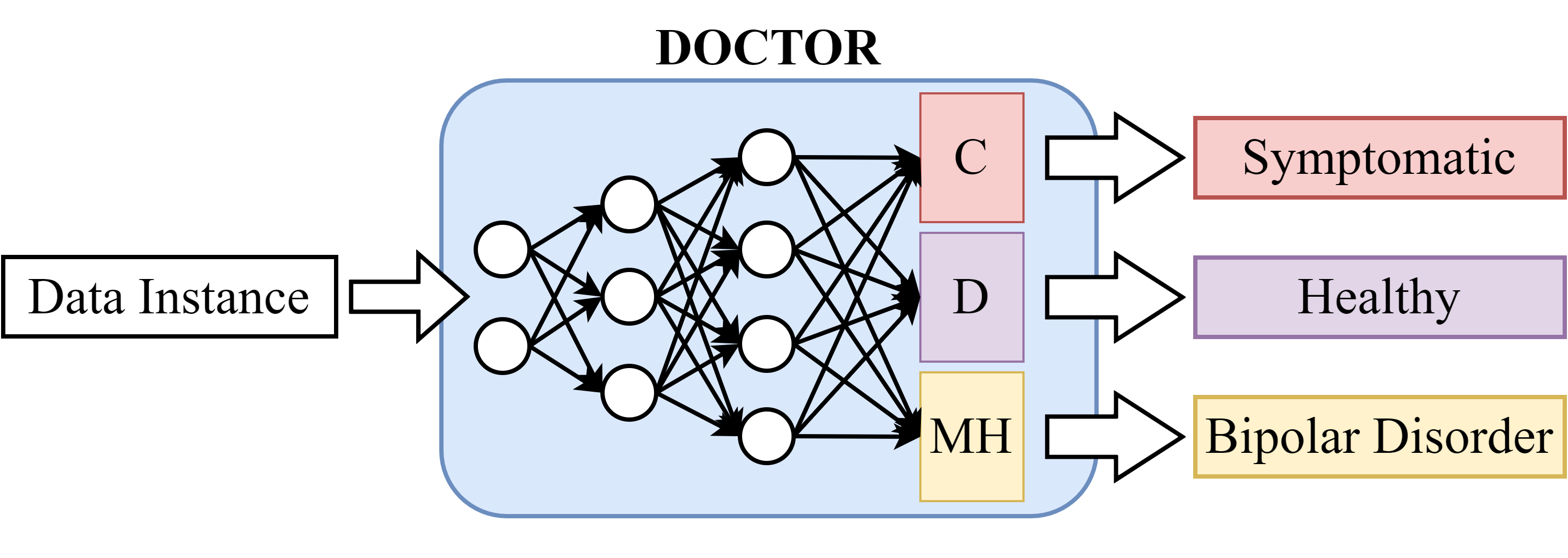}
    \caption{The demonstration of DOCTOR's simultaneous multi-disease detection application at test time. C represents the CovidDeep detection head. D symbolizes the DiabDeep detection head. MH stands for the MHDeep detection head.}
    \label{fig:detect}
\end{figure}

\subsection{Multi-disease Detection}
So far, we have validated DOCTOR's capacity in continually learning new missions in various CL 
scenarios through extensive experiments. Next, we discuss its prospects for simultaneous multi-disease detection.
Fig.~\ref{fig:detect} illustrates how DOCTOR can be used for simultaneous multi-disease detection at test time. 
For this demonstration, we use DOCTOR with the MLP model that has continually learned three disease detection 
tasks with the DP method described in Section \ref{sec:TIL}. Now the model is capable of detecting 
three different diseases through its three detection heads. We pass the WMS data of a patient with Type-I diabetes 
in the DiabDeep dataset to DOCTOR. It informs us that the patient is detected as healthy for COVID-19, 
Type-I diabetic, and positive for bipolar disorder. Hence, the multi-disease detection results can alert the 
patient to seek further medical advice from corresponding health providers.

Currently, we do not have access to datasets with labels associated with co-morbidities for each 
patient. Therefore, we are not able to conduct rigorous experiments for this application in this work. However, we 
have shown the promising potential of DOCTOR for this application. On the other hand, it is worth noting that 
DOCTOR employs DNNs to continually learn joint probability distributions of WMS data and their labels to perform 
disease detection. Thus, it is best to train DOCTOR with disease detection tasks that take as input WMS data 
containing the same physiological signals to obtain optimal multi-disease detection performance.

\subsection{Cost Analysis}
\label{sec:cost}

% This is a table made for pre-FPO
% \begin{table*}[t]
%     \caption{Compare Acc.}
%     \centering
%     \resizebox{0.55\linewidth}{!}{
%     \begin{tabular}{c|cc}
%     \toprule 
%     Frameworks & Test Accuracy & F1-score \\
%     \midrule
%     CovidDeep Model & 0.994 & 0.997 \\
%     DiabDeep Model & 0.945 & 0.946 \\
%     MHDeep Model & 0.865 & 0.976 \\
%     \bottomrule
%     \end{tabular}
%     }
%     \label{tbl:acc_compare}
% \end{table*}

One of the advantages of DOCTOR over conventional ML disease detection methods is model efficiency at 
test time. By utilizing the proposed replay-style CL algorithm, DOCTOR enables simultaneous multi-disease detection 
in users with a single multi-headed ML model. Conventional methods that do not employ CL strategies require separate 
models for different disease detection tasks. Hence, users have to install several models on their devices to detect 
different diseases. This increases storage and energy consumption.

Next, we perform a test-time cost analysis to detect the three distinct diseases. 
Table~\ref{tbl:cost} presents the analysis results on an arbitrary edge device for DOCTOR and 
conventional methods that do not employ CL strategies. Since the actual energy consumption would differ depending on 
the embedded system platform used, the number of floating point operations (FLOPs) is often used as a proxy for energy 
consumption \cite{flops1, flops2, flops3}. Therefore, we report the total number of models used, parameters stored, and FLOPs 
performed on an arbitrary edge device for evaluation. In addition, we report the total model size in MB. As shown 
in the table, only one model with roughly a third of parameters, FLOPs, and model size is required in DOCTOR 
relative to conventional methods.

Note that the CL training process for learning a new disease detection task should be
performed offline on a centralized server due to data privacy concerns. The updated model and data preprocessing protocol for 
inference can then be deployed on each user's edge device through a regular software update process. Therefore, in the 
preserved data method, the memory buffer, whose size is shown in Tables~\ref{tbl:DIL}, \ref{tbl:CIL}, and \ref{tbl:TIL}, 
is only needed on the central server. The memory storage required on the edge device is only to store the inference model.

\begin{table*}[t]
    \caption{Test-time Cost Analysis for Detecting the Three Diseases on an Arbitrary Edge Device}
    \centering
    \resizebox{0.8\linewidth}{!}{
    \begin{tabular}{c|cccc}
    \toprule 
    \multirow{2}{*}{Frameworks} & Number of & Total Number & Total Number & Total Model \\
     & Models & of Parameters & of FLOPs & Size (MB) \\
    \midrule
    Conventional Methods & \multirow{2}{*}{3} & \multirow{2}{*}{269,322} & \multirow{2}{*}{547.08k} & \multirow{2}{*}{1.05} \\
    (no CL Strategy) & & & & \\
    DOCTOR & 1 & 90,634 & 180.74k & 0.35 \\
    \bottomrule
    \end{tabular}
    }
    \label{tbl:cost}
\end{table*}

\subsection{Ablation Study}
\label{sec:ablation}

Next, we present results for several ablation studies.

\subsubsection{Informative Subset for Data Preservation}
\label{sec:ablation1}
Here, we report results for an ablation study on the selection of the most informative training data subset that 
should be preserved for the DP CL algorithm. We repeat the CL experiments described in Sections \ref{sec:DIL}, 
\ref{sec:CIL}, and \ref{sec:TIL} with only the DP method under the same settings for all three datasets. However, 
we modify the threshold value (\emph{i.e.}, the $p$-th percentile hyperparameter described in Section \ref{sec:DP}) 
to preserve different subsets of real training data. We repeat the experiments three times while preserving the 
real data instances whose average training loss values are above the 50th percentile (the top 50\%), between the 
75th and 25th percentiles (the middle 50\%), and below the 50th percentile (the bottom 50\%).

Table~\ref{tbl:ablation_subset} shows the experimental results of the ablation study. The best 
results are shown in bold. We can see that preserving the real training data subset whose average training loss 
values are above the 50th percentile (the top 50\%) results in the best performance ($Acc_{avg}$, 
F1-score$_{avg}$, and BWT) in all the experiments. These results demonstrate that training data with higher 
average training loss values are more informative. Hence, they can better assist the model in retaining the 
learned knowledge. Therefore, in our DP method, we preserve the real training data subset with 
the highest average training loss values.

\begin{table}[t]
    \caption{Ablation Study on Informative Subset for Data Preservation}
    \centering
    \resizebox{0.65\linewidth}{!}{
    \begin{tabular}{c|c|cccccc}
    \toprule 
    CL & \multirow{2}{*}{Datasets} & \multirow{2}{*}{Subset} & \multirow{2}{*}{$Acc_{avg}$} & \multirow{2}{*}{F1-score$_{avg}$} & \multirow{2}{*}{BWT} & Buffer Size\\
    Experiments & & & & & & (MB) \\
    \midrule
     & \multirow{3}{*}{CovidDeep} & Top 50\% & \textbf{0.991} & \textbf{0.997} & \textbf{-0.004} & 2.8 \\
     & & Middle 50\% & 0.977 & 0.984 & -0.033 & 2.8 \\
     & & Bottom 50\% & 0.970 & 0.979 & -0.046 & 2.8 \\
    \cmidrule{2-7}
    \multirow{2}{*}{Domain-} & \multirow{3}{*}{DiabDeep} & Top 50\% & \textbf{0.954} & \textbf{0.958} & \textbf{-0.001} & 5.7 \\
    \multirow{2}{*}{incremental} & & Middle 50\% & 0.946 & 0.950 & -0.018 & 5.7 \\
     & & Bottom 50\% & 0.928 & 0.934 & -0.053 & 5.7 \\
    \cmidrule{2-7}
     & \multirow{3}{*}{MHDeep} & Top 50\% & \textbf{0.871} & \textbf{0.972} & \textbf{-0.014} & 6.8 \\
     & & Middle 50\% & 0.866 & 0.970 & -0.025 & 6.8 \\
     & & Bottom 50\% & 0.866 & 0.967 & -0.043 & 6.8 \\
    \midrule
     & \multirow{3}{*}{CovidDeep} & Top 50\% & \textbf{0.983} & \textbf{0.984} & \textbf{-0.029} & 2.4 \\
     & & Middle 50\% & 0.958 & 0.967 & -0.073 & 2.4 \\
     & & Bottom 50\% & 0.955 & 0.964 & -0.081 & 2.4 \\
    \cmidrule{2-7}
    \multirow{2}{*}{Class-} & \multirow{3}{*}{DiabDeep} & Top 50\% & \textbf{0.906} & \textbf{0.917} & \textbf{-0.052} & 4.6 \\
    \multirow{2}{*}{incremental} & & Middle 50\% & 0.889 & 0.913 & -0.077 & 4.6 \\
     & & Bottom 50\% & 0.861 & 0.883 & -0.133 & 4.6 \\
    \cmidrule{2-7}
     & \multirow{3}{*}{MHDeep} & Top 50\% & \textbf{0.896} & \textbf{0.984} & \textbf{-0.046} & 4.2 \\
     & & Middle 50\% & 0.870 & 0.972 & -0.085 & 4.2 \\
     & & Bottom 50\% & 0.828 & 0.942 & -0.168 & 4.2 \\
    \midrule
     & MHDeep & \multirow{2}{*}{Top 50\%} & \multirow{2}{*}{\textbf{0.931}} & \multirow{2}{*}{\textbf{0.968}} & \multirow{2}{*}{\textbf{0.022}} & \multirow{2}{*}{10.5} \\
    \multirow{2}{*}{Task-} & $\uparrow$ & & & & \\
    \multirow{2}{*}{incremental} & DiabDeep & Middle 50\% & 0.894 & 0.953 & -0.038 & 10.5 \\
     & $\uparrow$ & \multirow{2}{*}{Bottom 50\%} & \multirow{2}{*}{0.885} & \multirow{2}{*}{0.939} & \multirow{2}{*}{-0.051} & \multirow{2}{*}{10.5} \\
     & CovidDeep & & & & \\
    \bottomrule
    \end{tabular}
    }
    \label{tbl:ablation_subset}
\end{table}

\begin{figure}[t]
    \centering
    \begin{subfigure}[b]{0.45\textwidth}
        \centering
        \includegraphics[width=\textwidth]{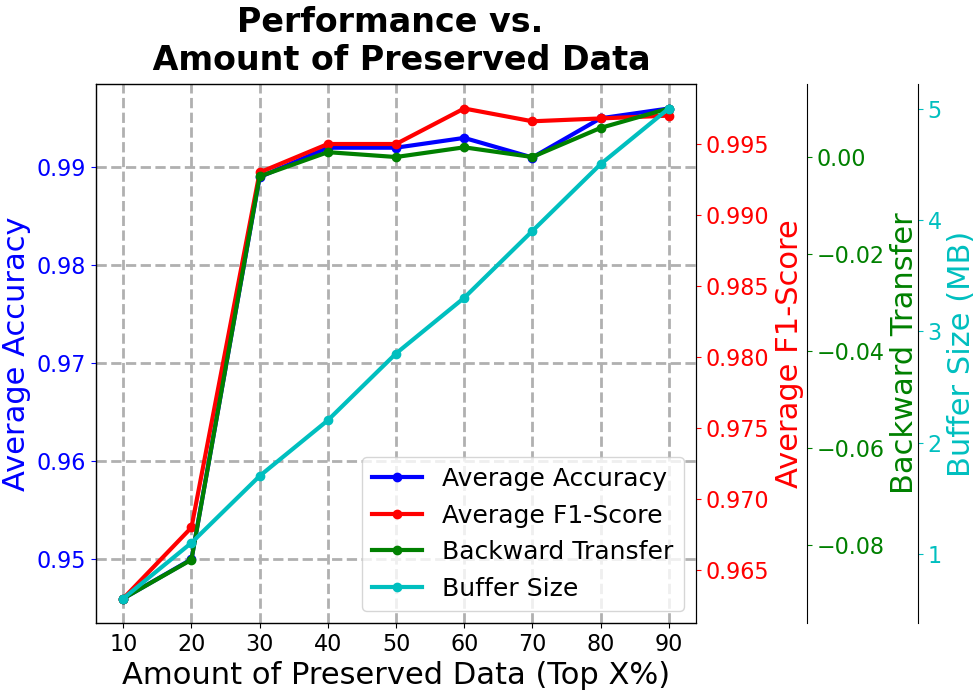}
        \caption{CovidDeep Dataset}
    \end{subfigure}
    \begin{subfigure}[b]{0.45\textwidth}
        \centering
        \includegraphics[width=\textwidth]{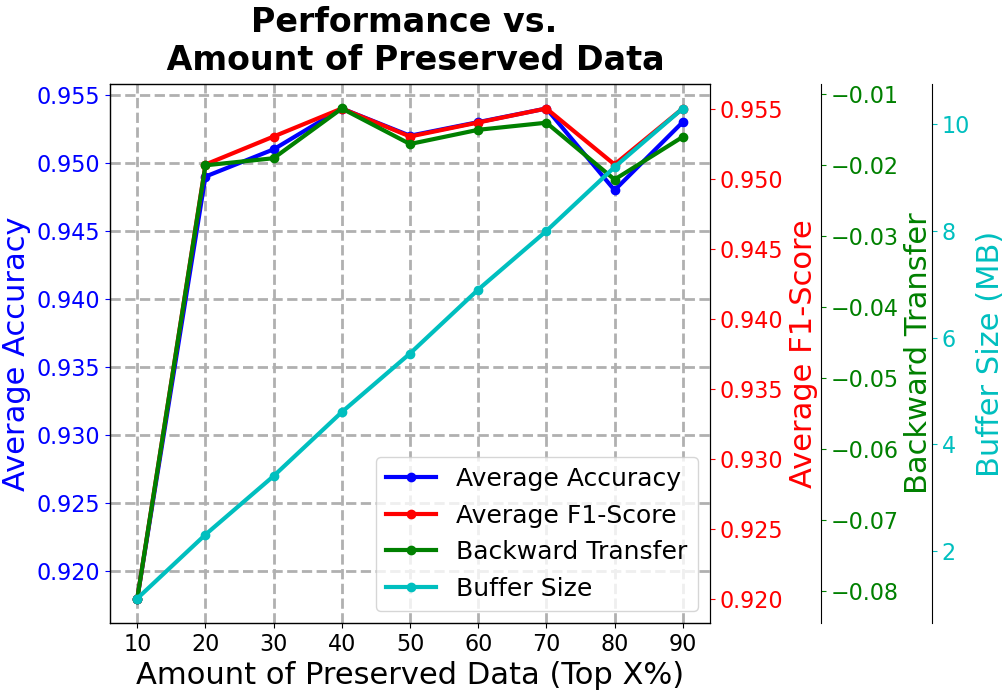}
        \caption{DiabDeep Dataset}
    \end{subfigure}
    \begin{subfigure}[b]{0.45\textwidth}
        \centering
        \includegraphics[width=\textwidth]{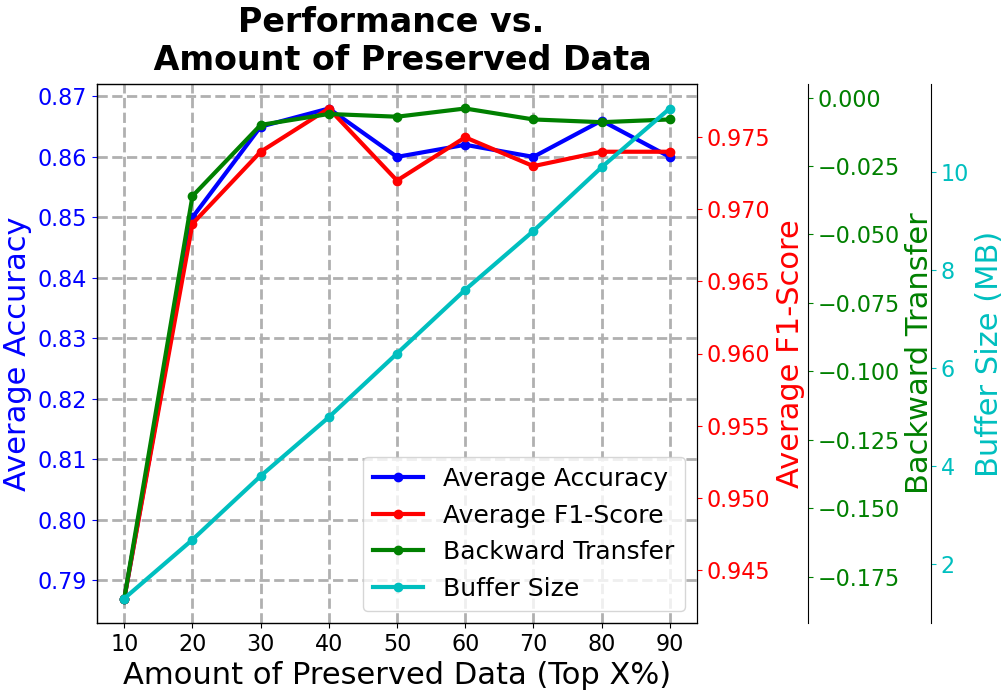}
        \caption{MHDeep Dataset}
    \end{subfigure}
    \caption{Performance vs.~the amount of real training data preserved in the \emph{domain-incremental} CL experiments for the (a) CovidDeep, (b) DiabDeep, and (c) MHDeep datasets. (Best viewed in color.)}
    \label{fig:abl_dp_DIL}
\end{figure}

\begin{figure}[t]
    \centering
    \begin{subfigure}[b]{0.45\textwidth}
        \centering
        \includegraphics[width=\textwidth]{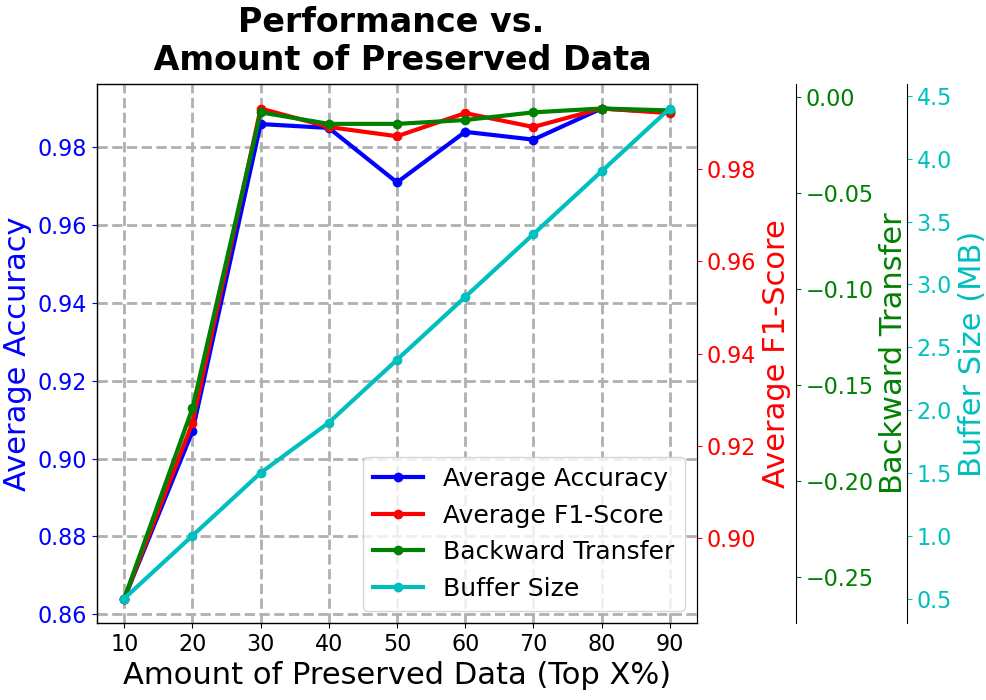}
        \caption{CovidDeep Dataset}
    \end{subfigure}
    \begin{subfigure}[b]{0.45\textwidth}
        \centering
        \includegraphics[width=\textwidth]{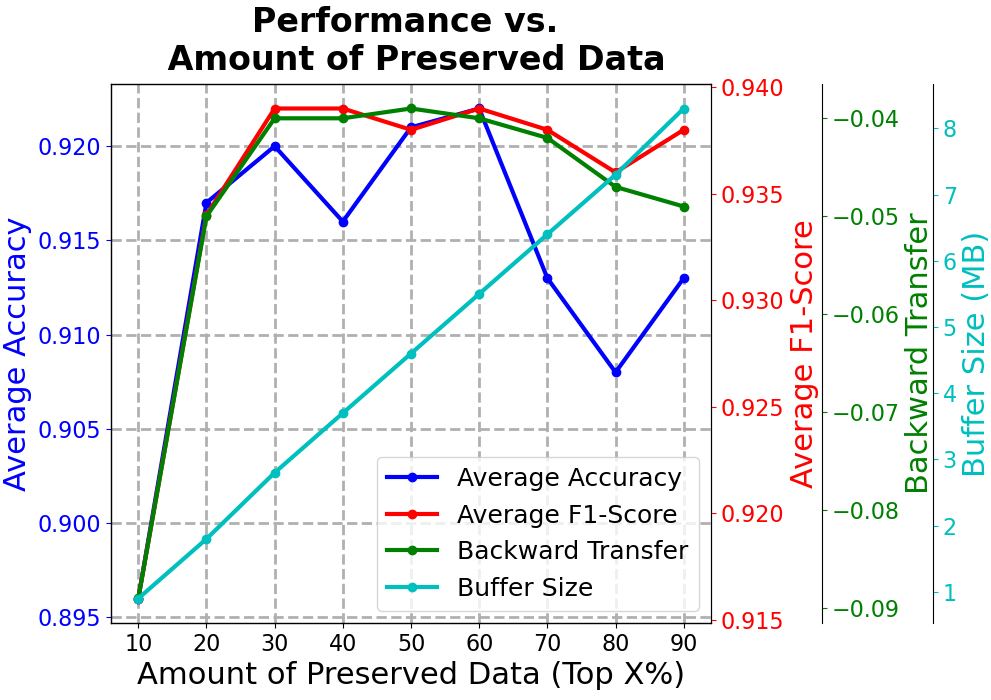}
        \caption{DiabDeep Dataset}
    \end{subfigure}
    \begin{subfigure}[b]{0.45\textwidth}
        \centering
        \includegraphics[width=\textwidth]{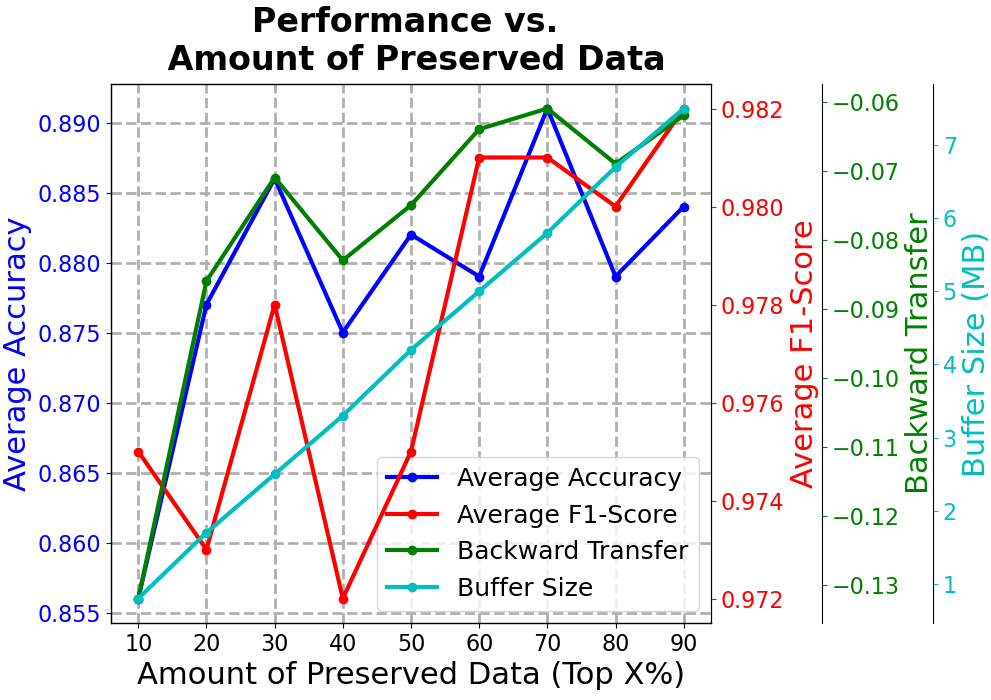}
        \caption{MHDeep Dataset}
    \end{subfigure}
    \caption{Performance vs.~the amount of real training data preserved in the \emph{class-incremental} CL experiments for the (a) CovidDeep, (b) DiabDeep, and (c) MHDeep datasets. (Best viewed in color.)}
    \label{fig:abl_dp_CIL}
\end{figure}

\begin{figure}[t]
    \centering    
    \includegraphics[width=0.5\linewidth]{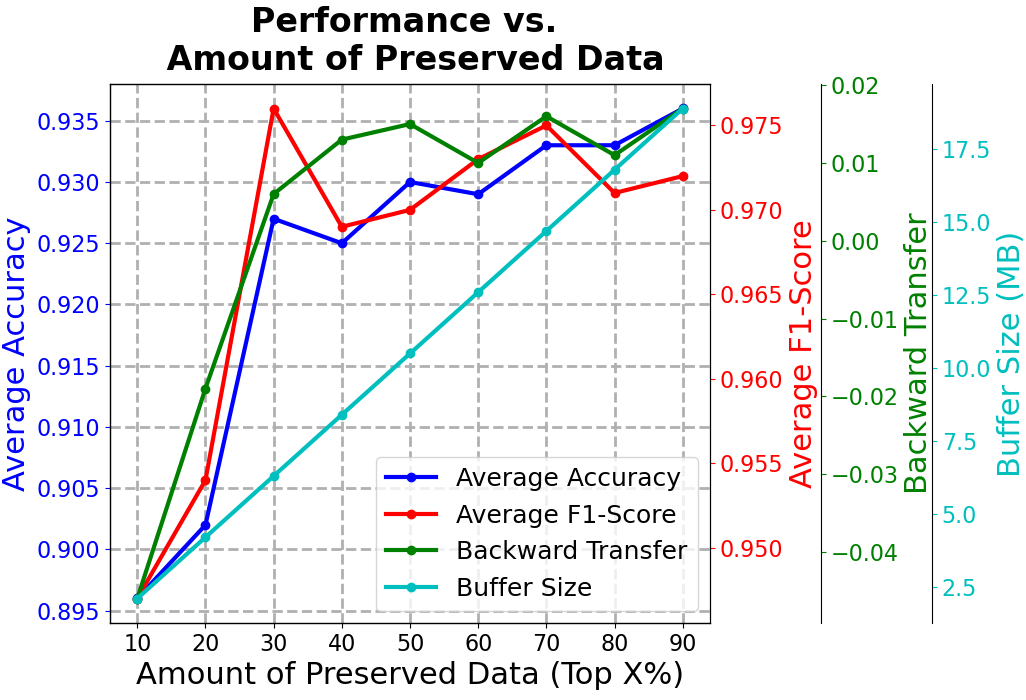}
    \caption{Performance vs.~the amount of real training data preserved in the \emph{task-incremental} CL 
experiments (CovidDeep $\rightarrow$ DiabDeep $\rightarrow$ MHDeep). (Best viewed in color.)}
    \label{fig:abl_dp_TIL}
\end{figure}

\subsubsection{Amount of Data Preserved}
\label{sec:ablation2}
Here, we perform an ablation study to examine the performance ($Acc_{avg}$, F1-score$_{avg}$, 
BWT, and buffer size) of the DP CL algorithm versus the amount of preserved real training data. We repeat the 
CL experiments again as in Section \ref{sec:ablation1} with only the DP method. However, this time, we repeat 
the experiments nine times while sweeping the threshold value from the 90th percentile to the 10th percentile 
in steps of 10 to preserve the top 10\%, top 20\%, $\dots$, and top 90\% of the real training data.

Fig.~\ref{fig:abl_dp_DIL}, \ref{fig:abl_dp_CIL}, and \ref{fig:abl_dp_TIL} show the experimental results 
of the ablation study in domain-, class-, and task-incremental CL experiments, respectively. This ablation study 
aims to find the optimal trade-off between the buffer size needed and other performance metrics. As we can see 
from the figures, preserving the top 30\% of real training data leads to minimum buffer requirements with 
acceptable performance loss. In general, preserving more real training data for replay should give better 
performance in CL. However, this does not always hold true. As the figures show, there might be a drop in 
$Acc_{avg}$, F1-score$_{avg}$, and BWT when we preserve more real training data. This is because the framework 
runs the risk of overfitting to previous missions and failing to learn the new ones when more preserved data are 
replayed. Moreover, preserving more data requires more memory storage and makes the framework harder to scale to 
many consecutive missions. Therefore, we set the threshold value to the 70th percentile to preserve the top 30\% 
of the real training data in the DP method.

\begin{figure}[t]
    \centering
    \begin{subfigure}[b]{0.85\textwidth}
        \centering
        \includegraphics[width=\textwidth]{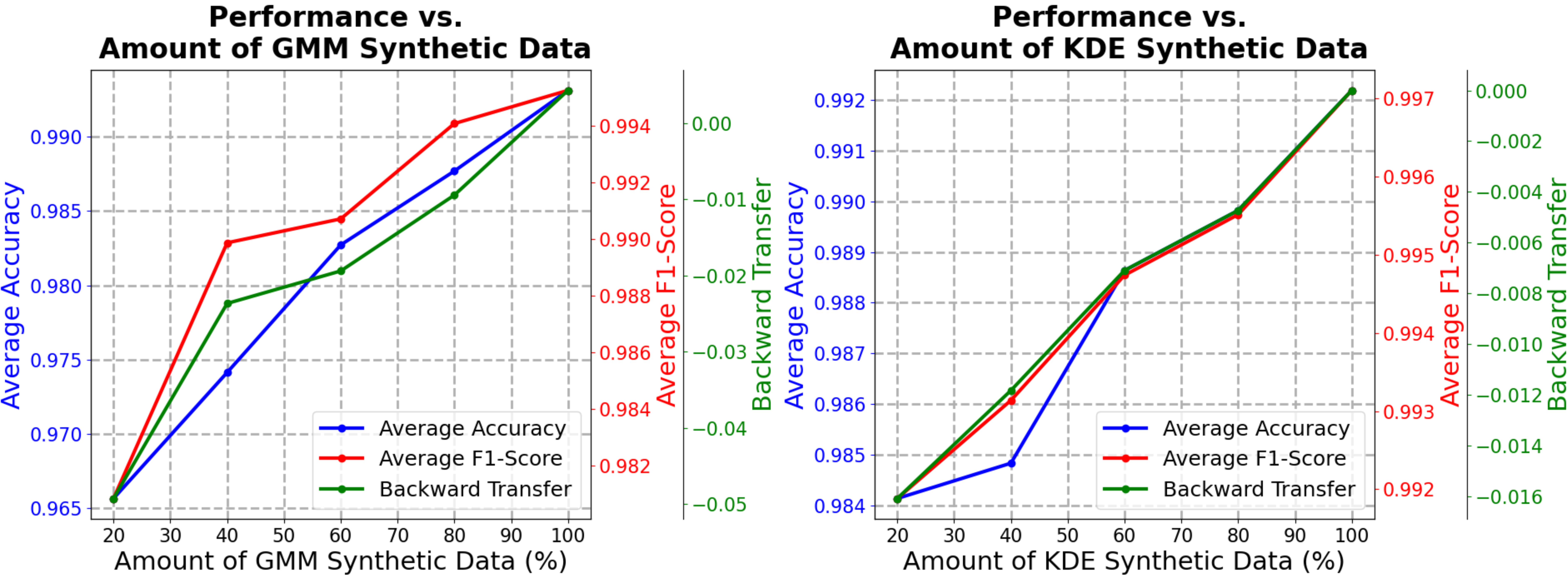}
        \caption{CovidDeep Dataset}
    \end{subfigure}
    \begin{subfigure}[b]{0.85\textwidth}
        \centering
        \includegraphics[width=\textwidth]{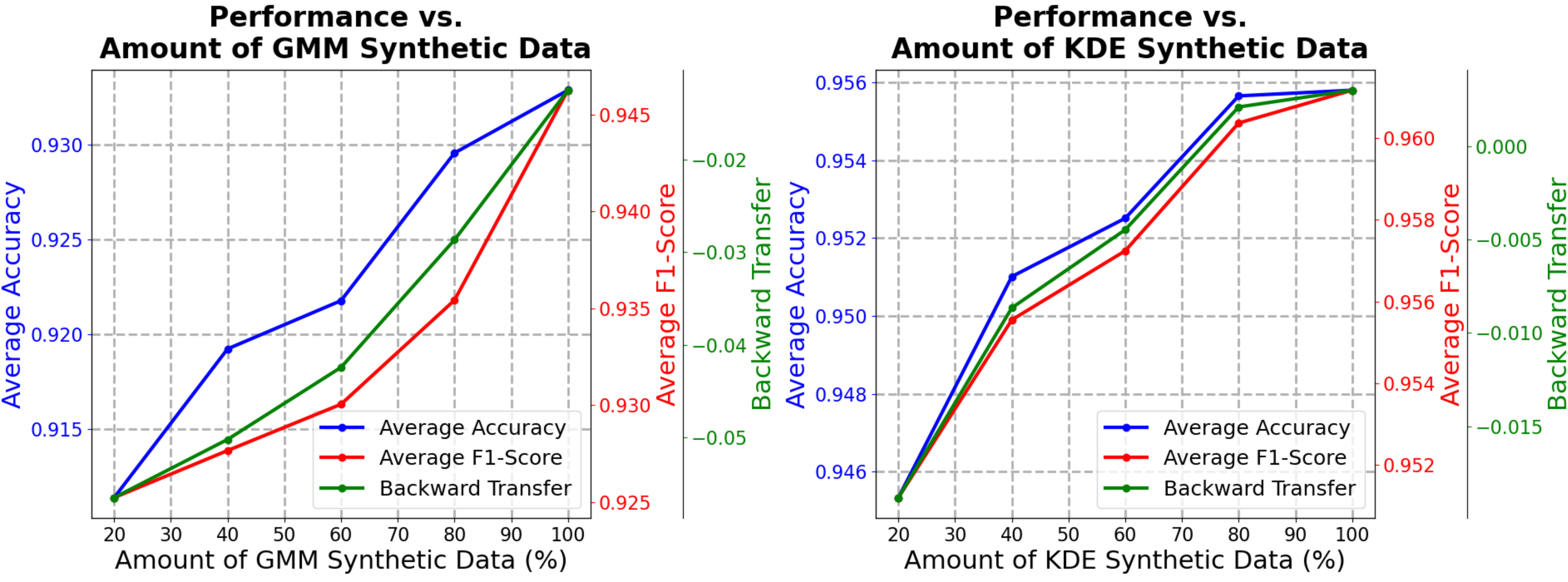}
        \caption{DiabDeep Dataset}
    \end{subfigure}
    \begin{subfigure}[b]{0.85\textwidth}
        \centering
        \includegraphics[width=\textwidth]{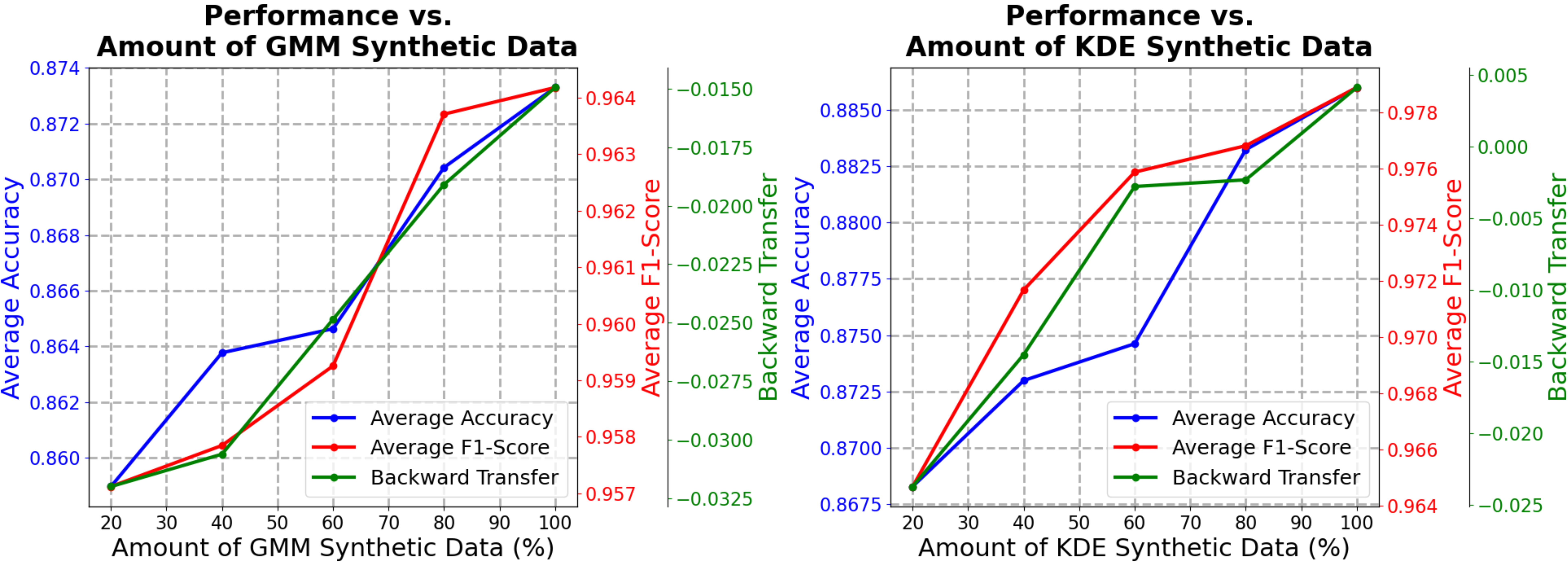}
        \caption{MHDeep Dataset}
    \end{subfigure}
    \caption{Performance vs.~the amount of synthetic data used in the \emph{domain-incremental} CL experiments for the (a) CovidDeep, (b) DiabDeep, and (c) MHDeep datasets. (Best viewed in color.)}
    \label{fig:abl_sdg_DIL}
\end{figure}

\begin{figure}[t]
    \centering
    \begin{subfigure}[b]{0.85\textwidth}
        \centering
        \includegraphics[width=\textwidth]{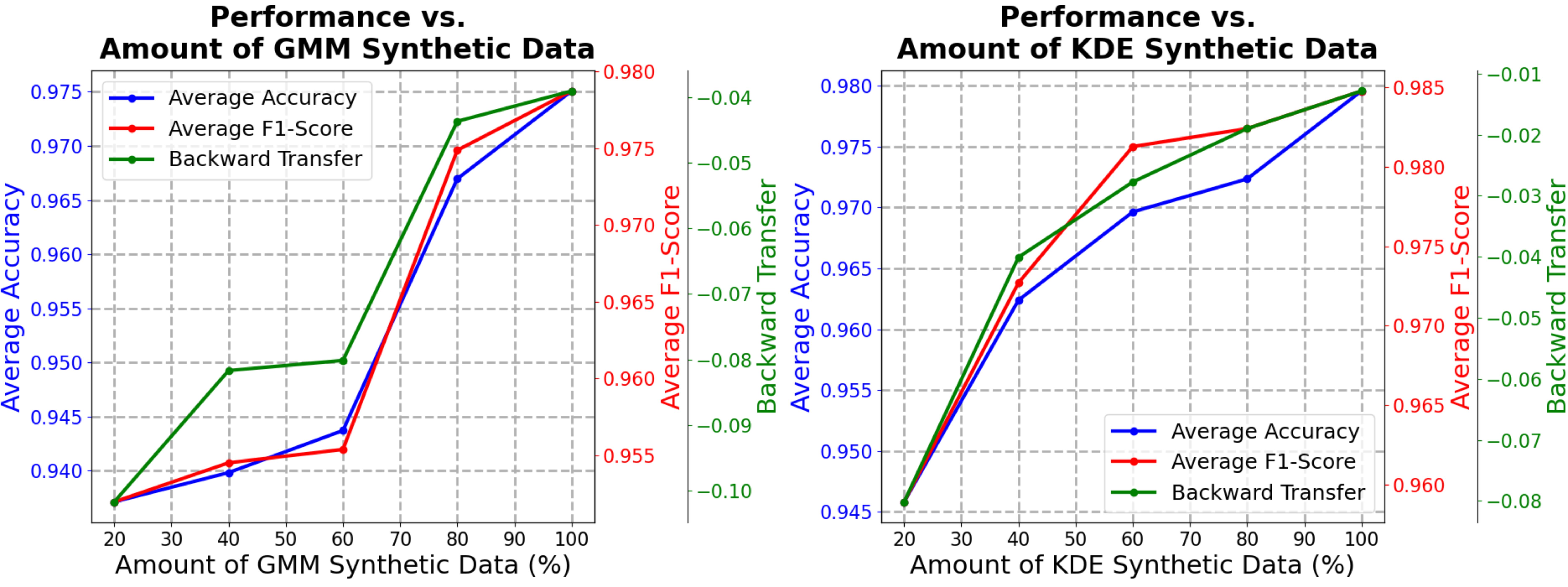}
        \caption{CovidDeep Dataset}
    \end{subfigure}
    \begin{subfigure}[b]{0.85\textwidth}
        \centering
        \includegraphics[width=\textwidth]{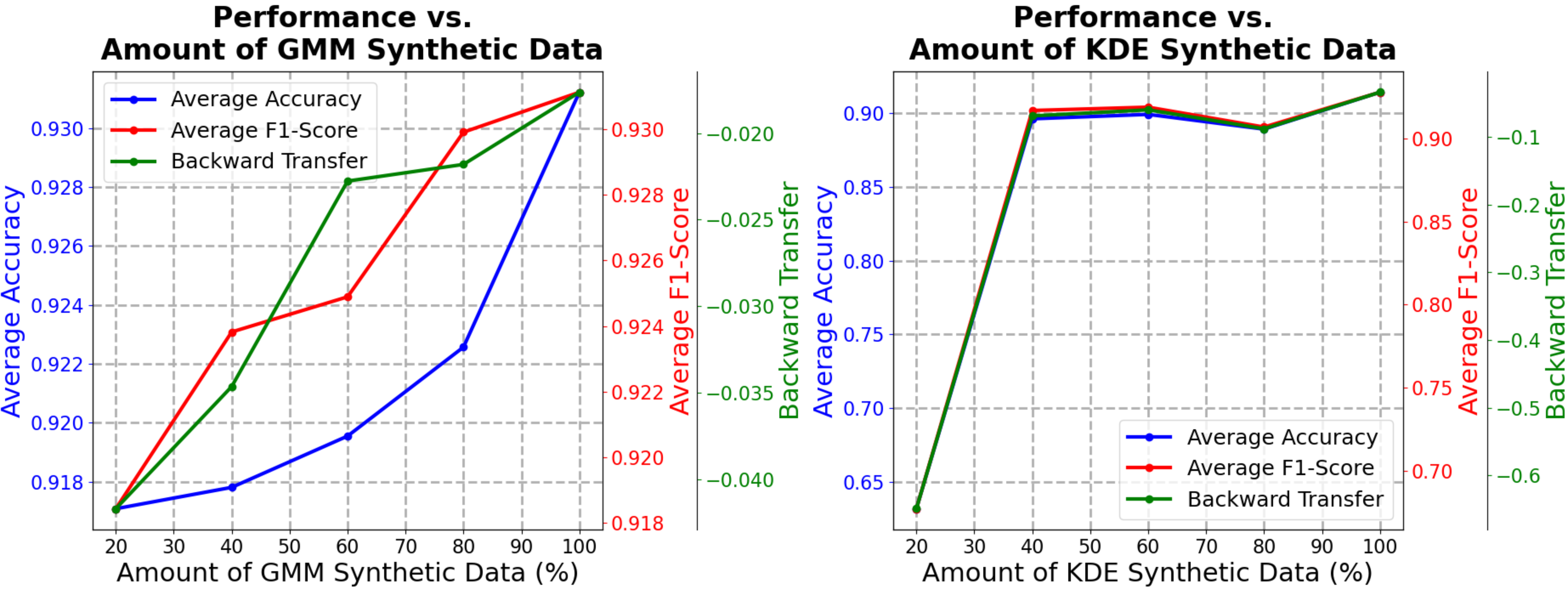}
        \caption{DiabDeep Dataset}
    \end{subfigure}
    \begin{subfigure}[b]{0.85\textwidth}
        \centering
        \includegraphics[width=\textwidth]{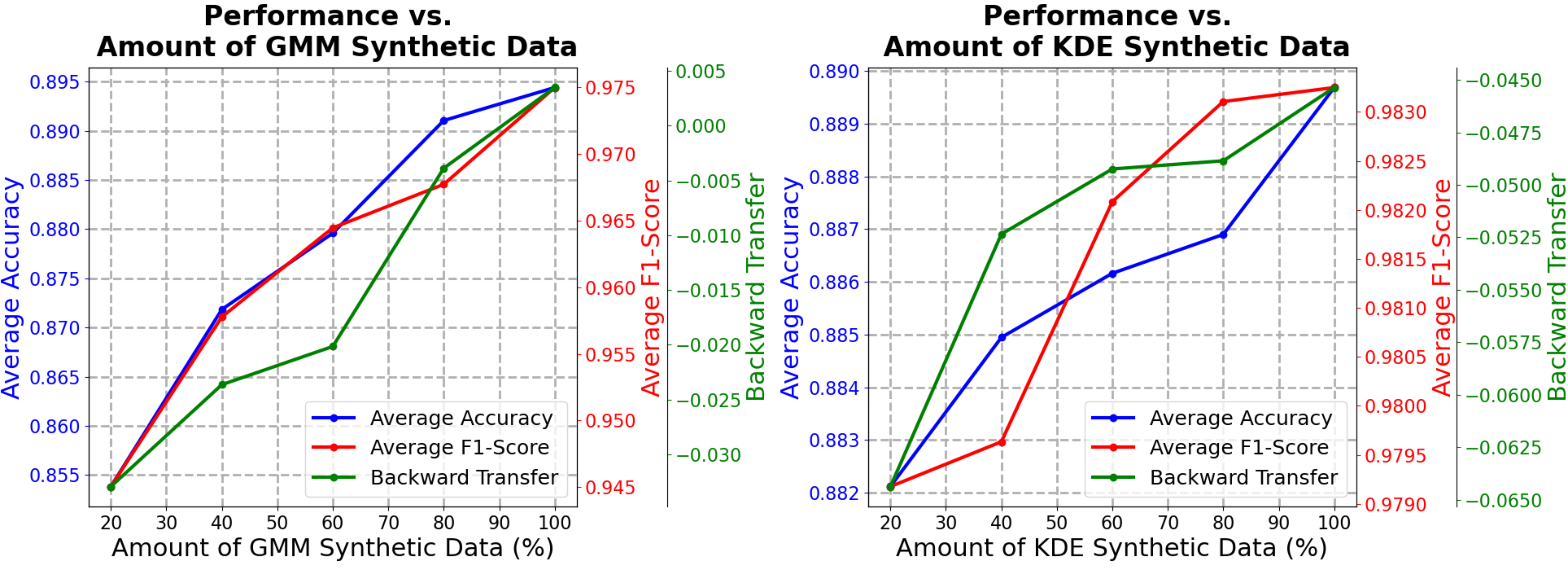}
        \caption{MHDeep Dataset}
    \end{subfigure}
    \caption{Performance vs.~the amount of synthetic data used in the \emph{class-incremental} CL experiments for the (a) CovidDeep, (b) DiabDeep, and (c) MHDeep datasets. (Best viewed in color.)}
    \label{fig:abl_sdg_CIL}
\end{figure}

\begin{figure}[t]
    \centering    
    \includegraphics[width=0.85\linewidth]{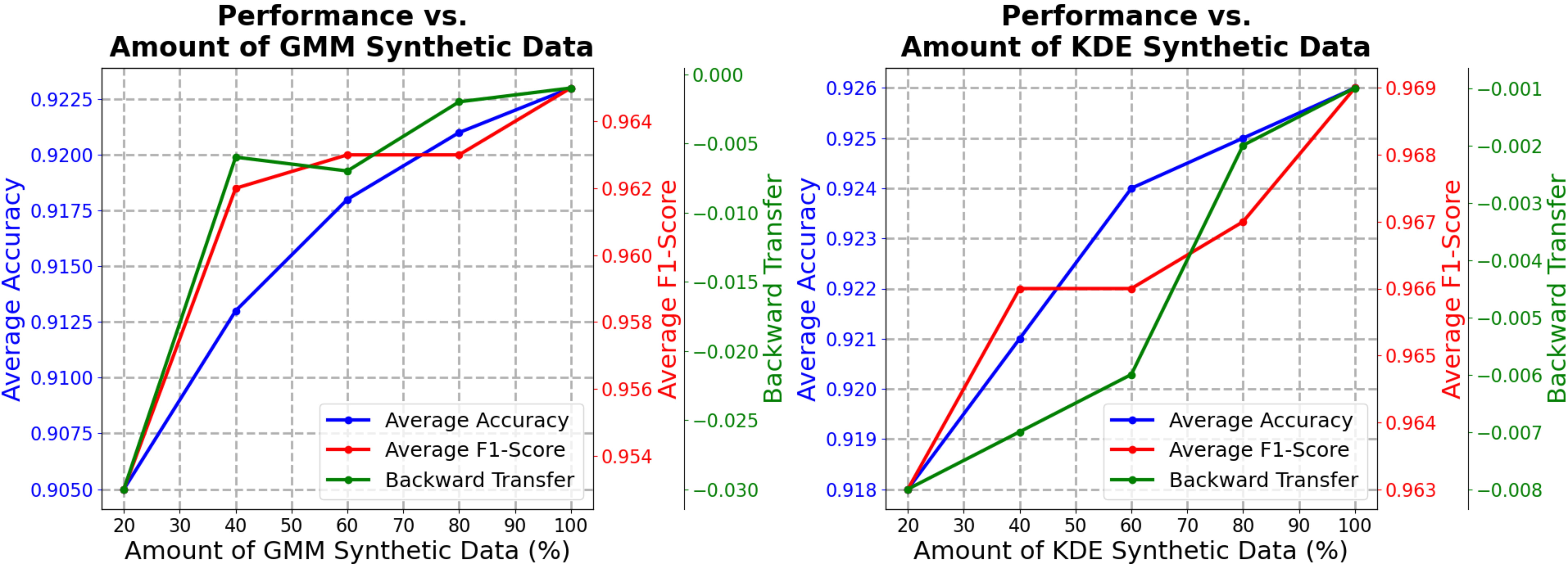}
    \caption{Performance vs.~the amount of synthetic data used in the \emph{task-incremental} CL 
experiments (CovidDeep $\rightarrow$ DiabDeep $\rightarrow$ MHDeep). (Best viewed in color.)}
    \label{fig:abl_sdg_TIL}
\end{figure}

\subsubsection{Amount of Synthetic Data}
\label{sec:ablation3}
Here, we report results for an ablation study of the performance ($Acc_{avg}$, F1-score$_{avg}$, 
and BWT) of the SDG CL algorithm versus the amount of synthetic data used. Since the SDG module generates synthetic 
data on the fly with the stored probability density estimation models during training, no extra memory buffer is 
needed. We repeat the experiments described in Sections \ref{sec:DIL}, \ref{sec:CIL}, and 
\ref{sec:TIL} with only the SDG method under the same settings for all three datasets. For the amount of synthetic
data generated, we target 20\%, 40\%, $\ldots$, and 100\% of the amount of real training data.

Fig.~\ref{fig:abl_sdg_DIL}, \ref{fig:abl_sdg_CIL}, and \ref{fig:abl_sdg_TIL} show the 
experimental results of the ablation study with both GMME and KDE methods in the SDG module in domain-, class-, and 
task-incremental CL experiments, respectively. The results demonstrate that more the synthetic data used in replay, 
better the performance in general. Since the SDG module generates synthetic data on the fly, the only constraint
is the capacity of the random-access memory (RAM) of the central server used for training. As long as the 
capacity allows, a designer can generate as much synthetic data as needed for replay.

\subsubsection{Alternative Machine Learning Model}
\label{sec:ablation4}
\begin{figure}[t]
    \centering
    \includegraphics[width=0.85\linewidth]{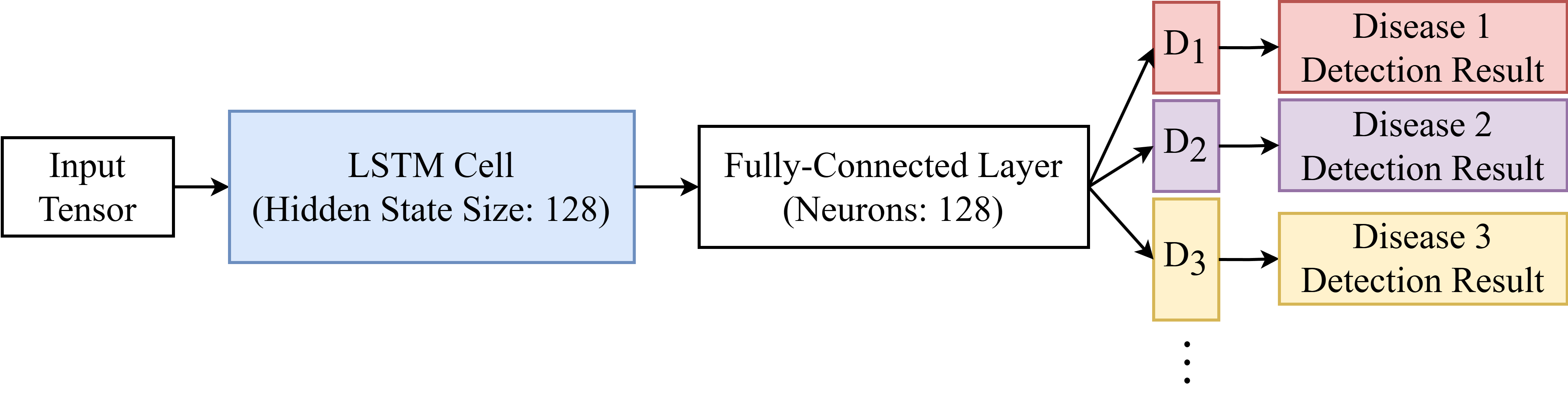}
    \caption{The LSTM network architecture in DOCTOR. D$_i$ represents the output head of the $i$-th disease detection task.}
    \label{fig:LSTM}
\end{figure}

Finally, we perform an ablation study to demonstrate DOCTOR's applicability to other
generic DNN models. We repeat the same domain-, class-, and task-incremental CL experiments described in 
Sections \ref{sec:DIL}, \ref{sec:CIL}, and \ref{sec:TIL}, respectively, with an LSTM network. We adopt the same datasets and preprocessing processes as described in Sections \ref{sec:datasets} and \ref{sec:preprocessing}. For each CL experiment, we follow the same procedures to prepare the training, validation, and test sets as mentioned in Sections \ref{sec:DIL}, \ref{sec:CIL}, and \ref{sec:TIL}. We perform a grid search over various hyperparameters on the validation sets of the three datasets to design the LSTM network. We find that a three-layer LSTM network performs the best in general. Fig. \ref{fig:LSTM} shows the architecture of the final LSTM model used in this ablation study. The first layer is a basic LSTM cell \cite{LSTM} with a hidden state size of 128. Then, it is followed by a fully-connected layer with 128 neurons and ReLU activation. Lastly, it incorporates a multi-headed softmax output layer. We implement this LSTM network in PyTorch for the baseline, naive fine-tuning, LwF, DOCTOR, and ideal joint-training frameworks in all CL scenarios for comparison. The LSTM network in our DOCTOR framework has 163,722 parameters and a size of 625KB. We apply the same training recipe to train all frameworks and adopt the same parameter settings for our DOCTOR framework as described in Section \ref{sec:implement}. Similarly, we perform the experiments on an NVIDIA A100 GPU with CUDA and cuDNN libraries to accelerate the experiments.

Table \ref{tbl:DIL_LSTM}, \ref{tbl:CIL_LSTM}, and \ref{tbl:TIL_LSTM} show the experimental results for this ablation study in domain-, class-, and task-incremental CL scenarios, respectively. The best DOCTOR results are shown in bold. For all datasets in all three CL scenarios, the naive fine-tuning framework still suffers from significant performance deterioration for previous missions due to catastrophic forgetting. On the other hand, despite using a different DNN, DOCTOR still greatly outperforms the naive fine-tuning and LwF frameworks in all experiments. Moreover, it achieves very competitive performance relative to the ideal joint-training framework in all CL scenarios. In fact, DOCTOR with the LSTM network even performs better than DOCTOR with the MLP model in the domain-incremental CL scenario, but at the cost of a larger model size. This ablation study demonstrates that DOCTOR is directly applicable to other DNN models and can still achieve high disease classification accuracy in various CL scenarios.

\begin{table*}[t]
    \caption{Domain-incremental CL Experimental Results (LSTM Network)}
    \centering
    \resizebox{1\linewidth}{!}{
    \begin{tabular}{crlccccccc}
    \toprule
    Datasets & \multicolumn{2}{c}{Frameworks} & $M_1~Acc$ & $M_2~Acc$ & $Acc_{avg}$ & F1-score$_{avg}$ & BWT & KS Test Statistic & Buffer Size (MB) \\
    \midrule
    \multirow{7}{*}{CovidDeep} & \multicolumn{2}{c}{Ideal Joint-training} & 0.982 & 0.983 & 0.983 & 0.995 & -0.002 & - & 5.8 \\
     & \multirow{3}{*}{DOCTOR} & - DP & 0.929 & 0.983 & 0.956 & 0.977 & -0.055 & - & 1.7 \\
     & & - SDG with GMME & 0.938 & 0.996 & 0.967 & 0.980 & -0.045 & 0.083 & 0 \\
     & & - SDG with KDE & 0.978 & 0.999 & \textbf{0.988} & \textbf{0.994} & \textbf{-0.006} & 0.079 & 0 \\
     & \multicolumn{2}{c}{LwF} & 0.897 & 0.999 & 0.948 & 0.964 & -0.085 & - & 0 \\
     & \multicolumn{2}{c}{Naive Fine-tuning} & 0.786 & 1.000 & 0.893 & 0.936 & -0.198 & - & 0 \\
     \cmidrule{2-10}
     & \multicolumn{2}{c}{Baseline} & 0.983 & 0.819 & 0.901 & 0.937 & - & - & 0 \\
    \midrule
    \multirow{7}{*}{DiabDeep} & \multicolumn{2}{c}{Ideal Joint-training} & 0.917 & 0.996 & 0.957 & 0.957 & 0.011 & - & 11.5 \\
     & \multirow{3}{*}{DOCTOR} & - DP & 0.915 & 0.996 & 0.956 & 0.956 & 0.008 & - & 3.4 \\
     & & - SDG with GMME & 0.896 & 0.957 & 0.927 & 0.928 & -0.011 & 0.215 & 0 \\
     & & - SDG with KDE & 0.932 & 0.999 & \textbf{0.965} & \textbf{0.967} & \textbf{0.024} & 0.050 & 0 \\
     & \multicolumn{2}{c}{LwF} & 0.810 & 0.989 & 0.900 & 0.905 & -0.131 & - & 0 \\
     & \multicolumn{2}{c}{Naive Fine-tuning} & 0.631 & 1.000 & 0.816 & 0.825 & -0.276 & - & 0 \\
     \cmidrule{2-10}
     & \multicolumn{2}{c}{Baseline} & 0.907 & 0.372 & 0.640 & 0.649 & - & - & 0 \\
    \midrule
    \multirow{7}{*}{MHDeep} & \multicolumn{2}{c}{Ideal Joint-training} & 0.831 & 0.953 & 0.892 & 0.974 & -0.012 & - & 13.6 \\
     & \multirow{3}{*}{DOCTOR} & - DP & 0.820 & 0.955 & \textbf{0.888} & \textbf{0.977} & -0.023 & - & 4.1 \\
     & & - SDG with GMME & 0.832 & 0.898 & 0.865 & 0.967 & \textbf{-0.011} & 0.041 & 0 \\
     & & - SDG with KDE & 0.832 & 0.924 & 0.878 & 0.967 & \textbf{-0.011} & 0.036 & 0 \\
     & \multicolumn{2}{c}{LwF} & 0.657 & 0.943 & 0.800 & 0.883 & -0.197 & - & 0 \\
     & \multicolumn{2}{c}{Naive Fine-tuning} & 0.420 & 0.983 & 0.702 & 0.825 & -0.423 & - & 0 \\
     \cmidrule{2-10}
     & \multicolumn{2}{c}{Baseline} & 0.843 & 0.114 & 0.479 & 0.755 & - & - & 0 \\
    \bottomrule
    \end{tabular}
    }
    \label{tbl:DIL_LSTM}
\end{table*}

\begin{table*}[t]
    \caption{Class-incremental CL Experimental Results (LSTM Network)}
    \centering
    \resizebox{1\linewidth}{!}{
    \begin{tabular}{crlccccccc}
    \toprule
    Datasets & \multicolumn{2}{c}{Frameworks} & $M_1~Acc$ & $M_2~Acc$ & $Acc_{avg}$ & F1-score$_{avg}$ & BWT & KS Test Statistic & Buffer Size (MB) \\
    \midrule
    \multirow{7}{*}{CovidDeep} & \multicolumn{2}{c}{Ideal Joint-training} & 0.978 & 0.992 & 0.985 & 0.990 & -0.020 & - & 5.0 \\
     & \multirow{3}{*}{DOCTOR} & - DP & 0.807 & 0.983 & 0.895 & 0.914 & -0.192 & - & 1.5 \\
     & & - SDG with GMME & 0.961 & 0.984 & 0.973 & 0.981 & -0.037 & 0.166 & 0 \\
     & & - SDG with KDE & 0.987 & 0.972 & \textbf{0.980} & \textbf{0.992} & \textbf{-0.011} & 0.081 & 0 \\
     & \multicolumn{2}{c}{LwF} & 0.107 & 1.000 & 0.553 & 0.837 & -0.892 & - & 0 \\
     & \multicolumn{2}{c}{Naive Fine-tuning} & 0 & 1.000 & 0.500 & 0.837 & -0.998 & - & 0 \\
     \cmidrule{2-10}
     & \multicolumn{2}{c}{Baseline} & 0.998 & 0 & 0.499 & 0.766 & - & - & 0 \\
    \midrule
    \multirow{7}{*}{DiabDeep} & \multicolumn{2}{c}{Ideal Joint-training} & 0.882 & 0.943 & 0.913 & 0.922 & -0.042 & - & 9.2 \\
     & \multirow{3}{*}{DOCTOR} & - DP & 0.896 & 0.970 & 0.933 & 0.945 & -0.023 & - & 2.8 \\
     & & - SDG with GMME & 0.924 & 0.912 & 0.918 & 0.921 & \textbf{0.005} & 0.063 & 0 \\
     & & - SDG with KDE & 0.919 & 0.970 & \textbf{0.944} & \textbf{0.953} & 0 & 0.211 & 0 \\
     & \multicolumn{2}{c}{LwF} & 0 & 1.000 & 0.500 & 0.647 & -0.958 & - & 0 \\
     & \multicolumn{2}{c}{Naive Fine-tuning} & 0 & 1.000 & 0.500 & 0.647 & -0.919 & - & 0 \\
     \cmidrule{2-10}
     & \multicolumn{2}{c}{Baseline} & 0.919 & 0 & 0.459 & 0.696 & - & - & 0 \\
    \midrule
    \multirow{7}{*}{MHDeep} & \multicolumn{2}{c}{Ideal Joint-training} & 0.892 & 0.874 & 0.883 & 0.983 & -0.029 & - & 8.3 \\
     & \multirow{3}{*}{DOCTOR} & - DP & 0.893 & 0.889 & 0.891 & 0.980 & -0.045 & - & 2.5 \\
     & & - SDG with GMME & 0.927 & 0.867 & 0.897 & 0.953 & \textbf{-0.012} & 0.042 & 0 \\
     & & - SDG with KDE & 0.907 & 0.889 & \textbf{0.898} & \textbf{0.981} & -0.031 & 0.209 & 0 \\
     & \multicolumn{2}{c}{LwF} & 0.242 & 0.849 & 0.545 & 0.772 & -0.645 & - & 0 \\
     & \multicolumn{2}{c}{Naive Fine-tuning} & 0 & 0.904 & 0.452 & 0.775 & -0.938 & - & 0 \\
     \cmidrule{2-10}
     & \multicolumn{2}{c}{Baseline} & 0.938 & 0 & 0.469 & 0.500 & - & - & 0 \\
    \bottomrule
    \end{tabular}
    }
    \label{tbl:CIL_LSTM}
\end{table*}

\begin{table*}[t]
    \caption{Task-incremental CL Experimental Results (LSTM Network)}
    \centering
    \resizebox{1\linewidth}{!}{
    \begin{tabular}{crlcccccccc}
    \toprule 
    Datasets & \multicolumn{2}{c}{Frameworks} & $M_1~Acc$ & $M_2~Acc$ & $M_3~Acc$ & $Acc_{avg}$ & F1-score$_{avg}$ & BWT & KS Test Statistic & Buffer Size (MB) \\
    \midrule
    MHDeep & \multicolumn{2}{c}{Ideal Joint-training} & 0.993 & 0.920 & 0.838 & 0.917 & 0.965 & 0.014 & - & 21.1 \\
    $\uparrow$ & \multirow{3}{*}{DOCTOR} & - DP & 0.973 & 0.935 & 0.868 & \textbf{0.925} & \textbf{0.968} & \textbf{0.009} & - & 6.3 \\
    \multirow{2}{*}{DiabDeep} & & - SDG with GMME & 0.970 & 0.918 & 0.861 & 0.916 & 0.959 & -0.007 & 0.036 & 0 \\
    \multirow{2}{*}{$\uparrow$} & & - SDG with KDE & 0.963 & 0.921 & 0.868 & 0.917 & 0.962 & -0.004 & 0.041 & 0 \\
     & \multicolumn{2}{c}{LwF} & 0.307 & 0.921 & 0.877 & 0.702 & 0.916 & -0.349 & - & 0 \\
    CovidDeep & \multicolumn{2}{c}{Naive Fine-tuning} & 0.879 & 0.575 & 0.862 & 0.772 & 0.847 & -0.220 & - & 0 \\
    \midrule
    \multirow{2}{*}{DiabDeep} & \multicolumn{2}{c}{Ideal Joint-training} & 0.992 & 0.899 & - & 0.946 & 0.949 & 0.007 & - & 7.3 \\
     & \multirow{3}{*}{DOCTOR} & - DP & 0.960 & 0.917 & - & 0.938 & 0.951 & -0.012 & - & 2.2 \\
    \multirow{2}{*}{$\uparrow$} & & - SDG with GMME & 0.972 & 0.930 & - & \textbf{0.951} & \textbf{0.957} & \textbf{0} & 0.081 & 0 \\
     & & - SDG with KDE & 0.966 & 0.919 & - & 0.943 & 0.952 & -0.006 & 0.079 & 0 \\
    \multirow{2}{*}{CovidDeep} & \multicolumn{2}{c}{LwF} & 0.307 & 0.948 & - & 0.628 & 0.892 & -0.671 & - & 0 \\
     & \multicolumn{2}{c}{Naive Fine-tuning} & 0.931 & 0.923 & - & 0.927 & 0.945 & -0.041 & - & 0 \\
    \midrule
     CovidDeep & \multicolumn{2}{c}{Baseline} & 0.972 & - & - & 0.972 & 0.985 & - & - & 0 \\
    \bottomrule
    \end{tabular}
    }
    \label{tbl:TIL_LSTM}
\end{table*}

\section{Discussions and Limitations}
\label{sec:dis_lim}
In this section, we discuss some features and limitations of our DOCTOR framework. DOCTOR targets disease detection application tasks based on tabular data collected from WMSs. The DP method can be applied to medical image data as well since it samples preserved data based on their average training loss values only. However, the SDG module can only learn the probability distributions of tabular data. To enable full functionality to learn disease detection tasks based on medical images, a generative model for images would need to be employed in the framework. Either generative adversarial networks \cite{dgr, gan} or diffusion probabilistic models \cite{diffusion, diff_med_survey} may be a promising choice on this front. 

Another limitation of DOCTOR also lies in the SDG module. Both GMME and KDE methods model the joint probability distributions of real training data and sample synthetic data from the learned distributions without being aware of stratification. Hence, the synthetic data might be imbalanced in the classification classes. This would cause the framework to recall previous missions with biased predictions. One possible solution would be to learn individual probability distribution for each classification class and sample an even amount of synthetic data from each distribution. However, it would complicate the training process and incur extra computational costs. We leave this as future work.

\section{Conclusion}
\label{sec:conclusion}
We proposed DOCTOR, a multi-disease detection CL framework based on WMSs. It can continually learn new disease detection missions and perform simultaneous multi-disease detection with a single multi-headed DNN model and a replay-style CL algorithm. The proposed replay-style CL algorithm employs a DP method and an SDG module. The DP method preserves the most informative subset of real training data from previous missions without incurring excessive computational costs. The SDG module can generate as much synthetic data as needed from the learned joint multivariate probability distributions of the real training data from previous missions. 
In our experiments, we showed that DOCTOR is able to continually learn new data distributions, 
classification classes, and even different disease detection tasks while retaining prior knowledge and maintaining 
a small model size. We demonstrated that DOCTOR significantly outperforms the naive fine-tuning and LwF frameworks 
and performs very competitively relative to the ideal joint-training framework in three distinct CL scenarios with 
three different disease datasets.

\begin{acks}
This work was supported by NSF under Grant No. CNS-1907381. We would like to thank Sayeri Lala for developing code to prepare some of the datasets (specifically mental health and diabetes) used in this study and for her assistance in clarifying details concerning the data format.
\end{acks}

\bibliographystyle{ACM-Reference-Format}
\bibliography{biblio}

\end{document}